\documentclass{article}

\usepackage[preprint]{main}

\usepackage[utf8]{inputenc} 
\usepackage[T1]{fontenc}    
\usepackage{hyperref}       
\usepackage{url}            
\usepackage{booktabs}       
\usepackage{amsfonts}       
\usepackage{nicefrac}       
\usepackage{microtype}      
\usepackage{xcolor}         
\usepackage{amsmath}
\usepackage{graphicx} 
\usepackage{threeparttable}
\usepackage{multirow}
\usepackage{mathtools}
\usepackage{tabularx}
\usepackage{color}
\usepackage{colortbl}
\usepackage{amsthm}
\usepackage{caption}
\usepackage{marvosym}
\usepackage{comment}
\usepackage{float}  
\usepackage{tcolorbox} 

\definecolor{darklavender}{rgb}{0.45, 0.31, 0.59}
\definecolor{darkviolet}{rgb}{0.58, 0.0, 0.83}

\title{MV-CoLight: Efficient Object Compositing with Consistent Lighting and Shadow Generation}
    
\author{
Kerui Ren$^{1,2}$ \quad
Jiayang Bai$^{3}$ \quad
Linning Xu$^{4}$ \quad 
Lihan Jiang$^{2,5}$ \quad \\ 
\textbf{Jiangmiao Pang}$^{2}$ \quad
\textbf{Mulin Yu}$^{2}\thanks{Corresponding author.}$ \quad
\textbf{Bo Dai}$^{6}\protect\footnotemark[1]$ \quad \\
{\small$^1$Shanghai Jiao Tong University, $^2$Shanghai Artificial Intelligence Laboratory,} \\
{\small$^3$Nanjing University, $^4$The Chinese University of Hong Kong,} \\
{\small$^5$University of Science and Technology of China, $^6$The University of Hong Kong} \\
}

\begin{document}

\maketitle
{
\begin{center}
    \centering
    \captionsetup{type=figure}
    \includegraphics[width=\textwidth]{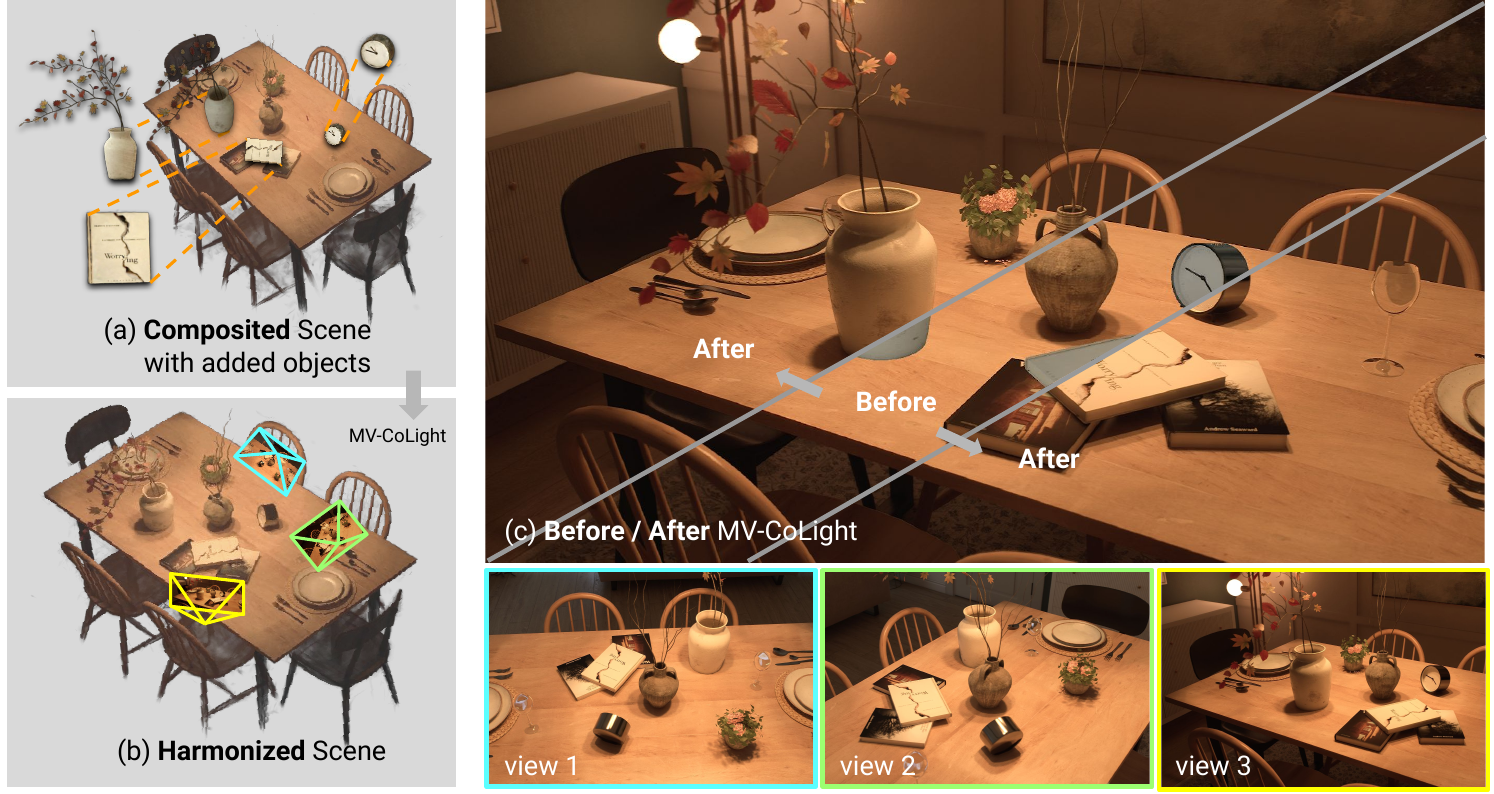}
    \captionof{figure}
    {
    Illustration of our object compositing pipeline with harmonization and relighting using MV-CoLight. In (a), we show a composite scene with visually inconsistent inserted objects. Applying our MV-CoLight method in (b), we generate realistic lighting, shadows, and harmonious integration of objects into the 3D scene. Panel (c) highlights clear visual differences before and after harmonization, accompanied by consistent novel view renderings below. Explore more demos on our project page: \href{https://city-super.github.io/mvcolight/}{\textcolor{magenta}{\textbf{https://city-super.github.io/mvcolight/}}}. 
    }    
\end{center}
}

\begin{abstract}
Object compositing offers significant promise for augmented reality (AR) and embodied intelligence applications. 
Existing approaches predominantly focus on single-image scenarios or intrinsic decomposition techniques, facing challenges with multi-view consistency, complex scenes, and diverse lighting conditions.
Recent inverse rendering advancements, such as 3D Gaussian and diffusion-based methods, have enhanced consistency but are limited by scalability, heavy data requirements, or prolonged reconstruction time per scene.
To broaden its applicability, we introduce MV-CoLight, a two-stage framework for illumination-consistent object compositing in both 2D images and 3D scenes. 
Our novel feed-forward architecture models lighting and shadows directly, avoiding the iterative biases of diffusion-based methods. We employ a Hilbert curve–based mapping to align 2D image inputs with 3D Gaussian scene representations seamlessly. To facilitate training and evaluation, we further introduce a large-scale 3D compositing dataset. 
Experiments demonstrate state-of-the-art harmonized results across standard benchmarks and our dataset, as well as casually captured real-world scenes demonstrate the framework’s robustness and wide generalization.
\end{abstract}
\section{Introduction}
\label{sec:intro}

Object compositing in 3D scenes remains a formidable challenge due to the interplay of color harmonization, shadow synthesis, light transport simulation, and multi‐view consistency, all of which must be addressed to achieve photorealistic integration. This capability is fundamental to AR, robotics, and interactive media, where realism directly impacts user immersion and perception.

Early object compositing research focuses primarily on isolated subtasks like scene relighting~\cite{iclight,neuralgaffer}, shadow generation~\cite{liu2020arshadowgan,liu2024shadow}, and color harmonization~\cite{cong2020dovenet,guerreiro2023pct}, yielding promising yet fragmented solutions.
However, The transition toward unified frameworks reveals intricate couplings between these components, necessitating adherence to physical principles governing light transport and occlusion phenomena.
Diffusion-based pipelines such as ObjectStitch~\cite{song2023objectstitch} and ControlCom~\cite{zhang2023controlcom} attempt single-image object insertion by synthesizing harmonious lighting and shadows within a background bounding box, but their reliance on stochastic sampling and the lack of large-scale, high-quality compositing datasets limit their robustness and generalization in real-world scenarios.

In this work, we tackle the problem of seamlessly inserting novel objects into static 3D scenes captured from multiple viewpoints. Our goal is to relight each object so that its appearance, including ambient illumination, surface reflections, and cast shadows, matches the lighting of the scene, while also modeling the reciprocal effects of the object on its surroundings (e.g. secondary shadows and interreflections). We introduce MV-CoLight, a unified framework that preserves both geometric fidelity and photorealism across views by learning and enforcing lighting‐consistent priors at both the image and scene levels.  
MV-CoLight adopts a two‐stage training pipeline. In the 2D object compositing stage, we train a feed‐forward model to capture scene-specific lighting characteristics, including background shadows and indirect illumination, from individual images. In the 3D object compositing stage, we transform these learned features into a 3D Gaussian representation using 3D Gaussian splatting~\cite{kerbl3Dgaussians}, ordering them via a Hilbert curve to ensure spatial coherence and enforce multi‐view consistency. 
Leveraging recent advances in video-level instance segmentation and 3D-aware object insertion, our framework effectively eliminates common 2D mask artifacts while achieving efficient inference (0.07s per frame) without compromising stability or visual quality.

To support training and evaluation, we introduce a large‐scale synthetic dataset of over 480k composite scenes rendered in Blender. Each scene features a table from the Digital Twin Catalog~\cite{dong2025digital}, augmented with Poly Haven HDR environment maps and materials~\cite{polyhaven}, and additional light sources for varied illumination. We render 16 uniformly sampled RGB views per scene, along with depth maps and segmentation masks. To simulate realistic compositing challenges, we mix foreground and background layers under different lighting conditions, creating deliberate lighting inconsistencies for training and evaluation. Further implementation details are provided in the supplementary material.  

Our main contributions are as follows:
1) a feed‐forward architecture for multi‐view object compositing that, unlike diffusion‐based alternatives, offers improved computational efficiency and robustness with high visual quality; 
2) a two-stage training framework that connects 2D object compositing with 3D Gaussian color fields via a Hilbert curve ordering mechanism, thereby enforcing geometrically consistent illumination priors and coherent multi‐view shadows; and 
3) curate a large‐scale benchmark of over 480 K annotated multi‐view scenes under varying lighting conditions, and demonstrate that our method achieves state‐of‐the‐art performance across several public datasets.
\section{Related Works}

Object compositing, the seamless integration of foreground objects into background scenes, is a fundamental task in both image editing and 3D graphics. In the following, we briefly discuss three principal paradigms that have guided existing solutions.

\noindent \textbf{Multi-Task Decomposition Approach.} 
Object compositing generally involves addressing three challenges, including color harmonization, relighting, and shadow generation. Below, we briefly review related works in these areas. Color harmonization has evolved from classical low-level techniques using color statistics and gradient adjustments~\cite{lalonde2007using, reinhard2001color, sunkavalli2010multi, wang2023semi} to learning-based methods~\cite{sofiiuk2021foreground, guo2021image, meng2025high, guerreiro2023pct, Guo_2021_CVPR, xue2022dccf, cong2022high} powered by large-scale datasets like iHarmony~\cite{cong2020dovenet}. Relighting modifies an object's shading while preserving its geometry and material properties. Recent learning-based relighting techniques focused on specific image types, including outdoor scenes~\cite{griffiths2022outcast, yu2020self}, portraits~\cite{iclight, rao2024lite2relight}, and human subjects~\cite{ji2022geometry, yoshihiro2018relighting}, achieving high-quality results. Shadow generation employs diverse strategies, from using pixel height information to generate diverse lighting effects~\cite{sheng2021ssn, sheng2022controllable} to GAN-based ~\cite{valencca2023shadow, zhang2019shadowgan, wang2024metashadowobjectcenteredshadowdetection} and generative models~\cite{liu2024shadow} that bypass ray-tracing requirements. While recent progress in these subdomains demonstrates improved fidelity, multi-view harmonization and physically grounded shadow synthesis remain open challenges, highlighting the need for holistic frameworks that ensure cross-task and cross-view coherence.

\noindent \textbf{End-to-End Unified Frameworks.}
Unified end-to-end frameworks for image compositing have emerged in recent studies~\cite{song2023objectstitch, chen2024mureobjectstitch, zhang2023controlcom, tarres2024thinking}. ObjectStitch~\cite{song2023objectstitch} introduces a diffusion-based architecture that concurrently addresses geometry correction, harmonization, shadow generation, and view synthesis. ControlCom~\cite{zhang2023controlcom} further enhances composite fidelity by incorporating a dedicated foreground refinement module. However, these approaches predominantly process single-view inputs. Building on ObjectStitch, MureObjectStitch~\cite{chen2024mureobjectstitch} adopts a multi-reference strategy for multi-perspective compositing, yet it still struggles with inconsistent harmonization when applied to multi-view images from the same scene. In contrast, our work leverages 3D modeling to ensure visual consistency across views, directly addressing these limitations. By integrating 3D priors, our approach simplifies the task to color-mapping transformations for inserted objects. This formulation inherently obviates the need for diffusion-based generative capabilities while necessitating precise per-pixel color transformations. Consequently, we employ a feed-forward network rather than diffusion-based models, which prioritize pixel-level generation and often yield unstable color outputs.

\noindent \textbf{Inverse Rendering Paradigm.} 
This approach for object compositing first estimates intrinsic scene properties, such as geometry, materials, and lighting, from input images through inverse rendering~\cite{barrow1978recovering}. Subsequently, traditional rendering pipelines or neural rendering pipelines are employed to render novel views of the scene with inserted objects. Recent advancements~\cite{jin2023tensoir, gao2024relightable, liang2024gs} have incorporated 3D scene representations like NeRF~\cite{mildenhall2021nerf} and 3D Gaussian Splats~\cite{kerbl3Dgaussians} within neural rendering pipelines. The emergence of large-scale image generative models~\cite{liang2024photorealistic, Zeng_2024, liang2025diffusionrenderer} has recently revitalized inverse rendering research. The RGB$\leftrightarrow$X~\cite{Zeng_2024} framework first trains an image diffusion model to estimate G-buffers from object and scene data. It then composites synthetic objects into these estimated channels and employs a diffusion model to generate final images with consistent lighting and shadow effects. However, such methods demand extensive high-quality datasets with fully paired intrinsic properties to achieve robust generalization capabilities, which poses significant challenges for real-world environment applications.

\section{Methods}
\label{sec:method}

\begin{figure*}[t!]
\centering
\includegraphics[width=\linewidth]{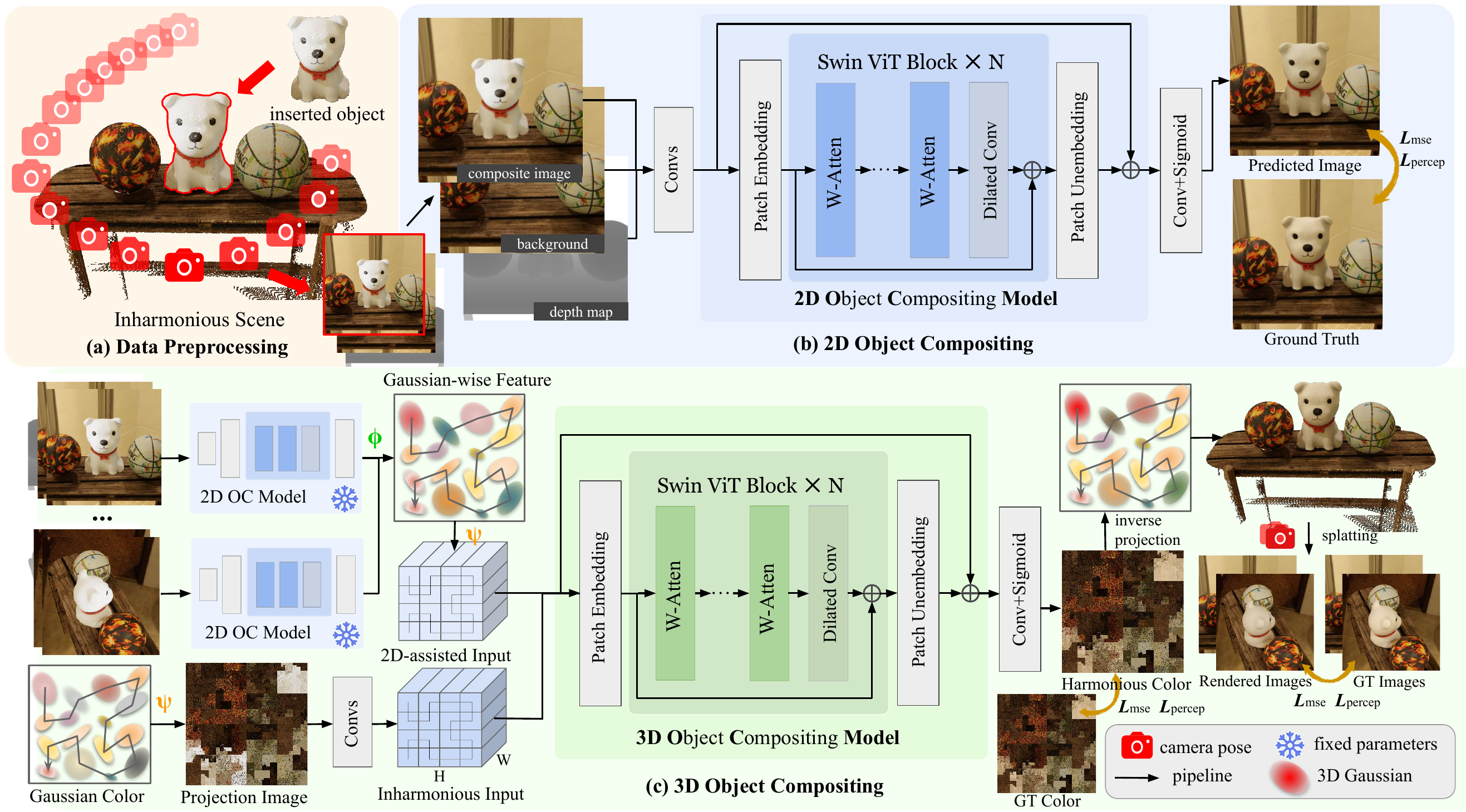}
\caption{Pipeline of MV-CoLight. In (a), we insert a white puppy as the composite object onto the table between basketballs, and render multi-view inharmonious images, background-only images, and depth maps using a camera trajectory moving from distant to close-up positions. Subsequently in (b), we input a single-view data into the 2D object compositing model, which processes the data through multiple Swin Transformer blocks to output the harmonized result. Finally in (c), we project the multi-view features from 2D models into Gaussian space via $\Phi(\cdot)$, combine them with the original inharmonious Gaussian colors projected into 2D Gaussian color space through $\Psi(\cdot)$, and then feed them into the 3D object compositing model. The model outputs harmonized Gaussian colors and computes rendering loss by incorporating Gaussian shape attributes.}
\label{fig:main-pipeline}
\end{figure*}

In this work, we focus on efficiently synthesizing consistent lighting and shadows to harmonize scenes. Formally, given a background scene and a foreground object, our task is to produce multi-view renderings that insert, harmonize, and relight the object under novel illumination while maintaining overall coherence, which presents an essential requirement for AR and embodied-intelligence applications that demand real-time, view-consistent integration. 

Fig.~\ref{fig:main-pipeline} illustrates our two‑stage framework. 
(1) We begin with an inharmonious scene and convert it into a pixel‑aligned 3D Gaussian representation (Sec.\ref{sec:data_preprocess}). 
(2) Each image is then processed independently by a transformer‑based network for single‑view object compositing (Sec.\ref{sec:2doc}). 
(3) To achieve multi‑view consistency, we concatenate the extracted pixel‑wise features with Gaussian‑wise features, order them along a Hilbert curve~\cite{hilbertcurve}, and decode them with a second transformer to predict harmonious Gaussian color attributes (Sec.\ref{sec:3doc}). 
(4) Finally, Sec.\ref{sec:loss} describes the loss functions that drive our two-stage training.
A brief introduction to Gaussian splatting and the Hilbert curve is provided in the supplementary materials.

\subsection{Data Preprocessing}
\label{sec:data_preprocess}
\begin{figure*}[t]
\centering
\includegraphics[width=\linewidth]{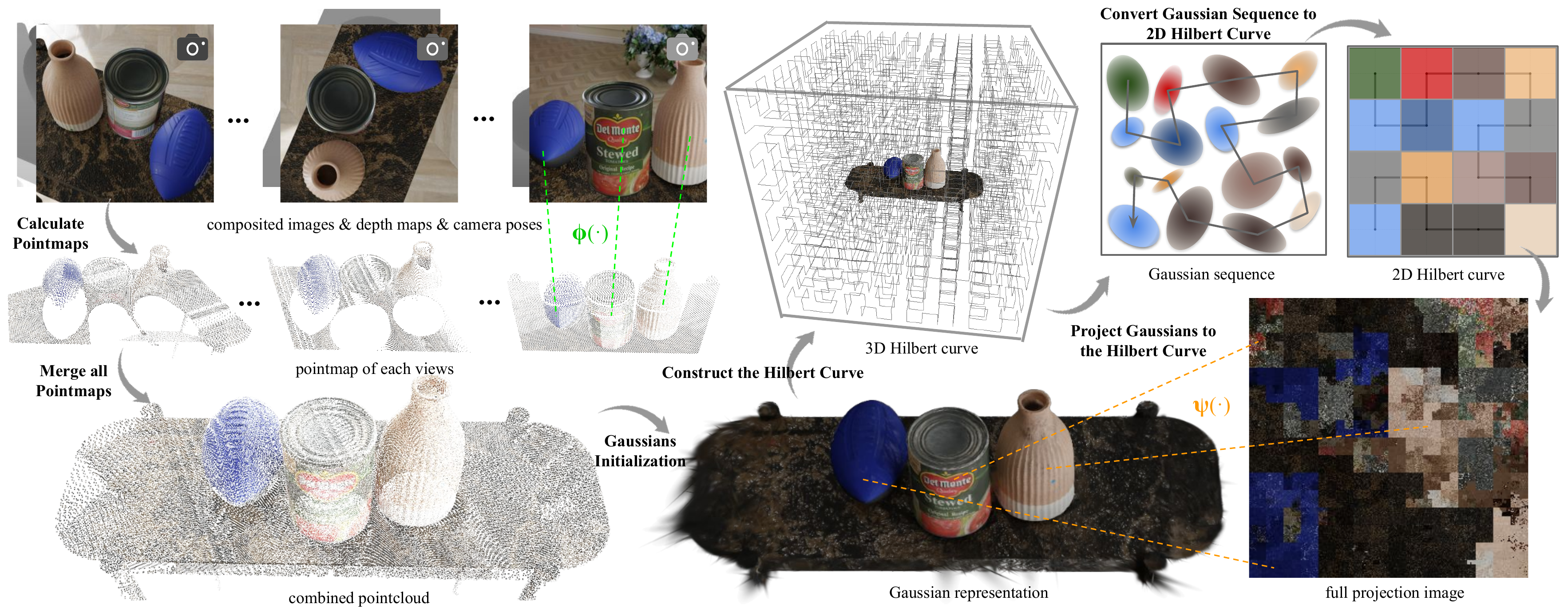}
\vspace{-1em}
\caption{
Mapping multi-view observations into a 2D Hilbert-ordered Gaussian color map. Starting from inharmonious multi-view images, depth maps, and camera poses, we compute per-view point maps and randomly sample $M$ points to initialize 3D Gaussian primitives, which we then optimize to fit the scene. Next, we construct a 3D Hilbert curve through the Gaussian centers and assign each primitive to its nearest curve point, yielding an ordered 1D sequence. Finally, we fold this sequence into a 2D grid along a 2D Hilbert curve, producing a spatially coherent projection in which each pixel encodes the color of its corresponding Gaussian.
}
\label{fig:hilbert}
\end{figure*}

Begin with a composed 3D scene, obtained via synthesis, 3D scanning, or multiview reconstruction pipeline (e.g., \cite{schoenberger2016sfm,wang2025vggt}), we place a set of cameras orbiting the scene center to obtain multi-view composite images, background-only images, and depth maps.
From each view’s images, camera poses, and depth data, we build point maps including 3D positions and colors, then randomly sample a fixed number $M$ of points to initialize the 3D Gaussian model $\mathcal{G}'$.
During optimization, we fix each Gaussian’s opacity at 1 and adjust only its shape parameters. 
As illustrated in Fig.~\ref{fig:hilbert}, we organize 3D Gaussian primitives, each tied to a unique training pixel, into spatially coherent patches by mapping their centers along a space-filling Hilbert curve\cite{hilbertcurve}, denoted as mapping $\Phi(\cdot)$.

\subsection{2D Object Compositing}
\label{sec:2doc}

Given an inharmonious composite image \( I \in \mathbb{R}^{3 \times H \times W} \), its background reference \( G \in \mathbb{R}^{3 \times H \times W} \), and depth map \( D \in \mathbb{R}^{1 \times H \times W} \), 
we form the input tensor $\{I, G, D\}$ and feed it into our 2D object compositing network:
\begin{equation} 
\hat{H} = \mathcal{M}_{2d}(\{I, G, D\}; \theta_{2d})
\label{eq:2doc}
\end{equation}
where $\hat{H}$ is the predicted harmonized image and $\theta_{2d}$ are the network parameters.

\noindent Specifically, \(\mathcal{M}_{2d}\) begins with several \(3\times3\) convolutions to extract shallow features, which are then partitioned into non-overlapping \(P\times P\) patches and fed into \(L\) Swin Transformer layers~\cite{liu2021swin}. For each layer \(i\in\{1,\dots,L\}\), the input \(F_{i-1}\) is normalized, processed by window-based multi-head self-attention with a residual connection, renormalized, passed through an MLP with a second residual skip, and output as \(F_i\), as summarized by
\begin{equation}
\label{eq:win_atten}
\begin{aligned}
\hat{F}_i &= \mathrm{W\mbox{-}Atten}\bigl(\mathrm{LN}(F_{i-1})\bigr) + F_{i-1},\\
F_i       &= \mathrm{MLP}\bigl(\mathrm{LN}(\hat{F}_i)\bigr) + \hat{F}_i.
\end{aligned}
\end{equation}

At inference time, $\mathcal{M}_{2d}$ is applied independently to each input. 
We extract the output feature maps ${F} \in \mathbb{R}^{m \times n \times H \times W}$ from the final attention block, where $m$ is the number of views and $n$ is the feature dimension.
These features are then transformed into 3D Gaussians via the mapping function $\Phi(\cdot)$ for downstream 3D compositing network (Sec.~\ref{sec:3doc}). See supplementary material for more details.

\subsection{3D Object Compositing}
\label{sec:3doc}

For 3D object compositing, we seek an illumination‐consistent Gaussian model \(\mathcal{G}\) that preserves each primitive’s 3D position \((x,y,z)\), scale, and rotation from the inharmonious model \(\mathcal{G}'\), but updates only its color attributes \(\mathcal{C'}\). Thus, we freeze all positional and geometric parameters and learn a color‐only mapping. To exploit efficient image‐transformer architectures, we use a mapping \(\Psi\) that first linearizes the sparse 3D Gaussians into a 1D sequence via a 3D Hilbert curve and then arranges them into a 2D grid by the inverse 2D Hilbert curve. We concatenate these Gaussian‐wise features, combining the transformed 2D colors and the \(\mathcal{M}_{2d}\) features via \(\Phi(\cdot)\), and feed them into our 3D compositing network:
\[
I_{\hat{\mathcal{C}}} \;=\;\mathcal{M}_{3d}\bigl(\bigl\{\Psi(\Phi(F)),\,\Psi(\mathcal{C}')\bigr\};\,\theta_{3d}\bigr),
\]
where \(\theta_{3d}\) are the network parameters and \(\mathcal{C}'\) are the original inharmonious Gaussian colors. The output \(I_{\hat{\mathcal{C}}}\) gives harmonious, view‐consistent colors, which we back‐project onto the 3D Gaussians to complete the compositing. 
Notably, \(\mathcal{M}_{3d}\) adopts the similar architectural designs as \(\mathcal{M}_{2d}\), differing only in its input and output dimensions. Please refer to supplementary material for more details.

\subsection{Loss Design}
\label{sec:loss}

During the two-stage training process, we employ similar loss function design, utilizing mean-square error loss $\mathcal{L}_{mse}$ loss and perceptual loss $\mathcal{L}_p$ to optimize the object compositing models:
\begin{equation}
\mathcal{L}_{2d}=\mathcal{L}_{mse}(\hat{H}, H)+\lambda \mathcal{L}_p(\hat{H}, H)
\end{equation}
\begin{equation}
\mathcal{L}_{3d}=\beta(\mathcal{L}_{mse}(I_{\hat{\mathcal{C}}}, \Psi(\mathcal{C}))+\lambda \mathcal{L}_{p}(I_{\hat{\mathcal{C}}}, \Psi(\mathcal{C})))+\frac{(1-\beta)}{n}\sum_{i=1}^{n}(\mathcal{L}_{mse}(\hat{H_{i}}, H_i)+\lambda \mathcal{L}_{p}(\hat{H_{i}}, H_i))
\end{equation}
where $H_{i}$ and $\hat{H_{i}}$ denote the ground truth images and the rendered images from the harmonized Gaussian $\mathcal{G'}$, which is composed of $\mathcal{C'}$ and shape parameter from $\mathcal{G}$, $\lambda$ and $\beta$ are the hyper-parameter and set as $0.05$ and $0.5$ by default.
\section{Experiments}
\label{sec:exp}

\subsection{Experimental Setup}
\paragraph{Datasets and Metrics.}
We evaluate our method on two public benchmarks, FOSCom~\cite{zhang2023controlcom} and Objects With Lighting (OWL)~\cite{ummenhofer2024objects}as well as on our newly curated dataset. For 2D object compositing, we test on 640 scenes from FOSCom, 72 scenes from OWL, and 57 challenging scenes from our proposed dataset. For 3D object compositing, we report results on 50 simple and 7 complex synthetic scenes from our dataset, with another two real captured scenes. All images are center-cropped and rescaled to 
$256 \times 256$ for uniform comparison. 
Performance is quantified using PSNR, SSIM~\cite{wang2004image}, and LPIPS~\cite{zhang2018unreasonable}. Since each ground-truth image embodies only one physically plausible lighting/albedo configuration, perceptual metrics (SSIM and LPIPS) offer additional assessments of structural and visual fidelity than PSNR alone.

\begin{table*}[t!]
\renewcommand{\arraystretch}{1.15}
\setlength{\tabcolsep}{3pt}
\centering
\vspace{-0.5em}
\caption{Single-view quantitative performance on our purposed dataset and the Objects With Lighting dataset~\cite{ummenhofer2024objects}. We report visual quality metrics, inference time and memory storage, highlighting the \textbf{best} and \underline{second-best} in each category. Our* and Our$\dag$ denote our method without depth input and without both depth and background input, respectively.}
\resizebox{1\linewidth}{!}{
\begin{tabular}{c|c|ccc|ccc|ccc|c|c}
\toprule
\multicolumn{2}{c|}{Dataset} &  \multicolumn{3}{c|}{Simple Synthetic Scene} & \multicolumn{3}{c|}{Complex Synthetic Scene} & \multicolumn{3}{c|}{Objects With Lighting} & \multirow{2}{*}{Time\(\downarrow\)}  & \multirow{2}{*}{Memory\(\downarrow\)}  \\ 
Paradigm & Method & PSNR\(\uparrow\) & SSIM\(\uparrow\) & LPIPS\(\downarrow\) & PSNR\(\uparrow\) & SSIM\(\uparrow\) & LPIPS\(\downarrow\) & PSNR\(\uparrow\) & SSIM\(\uparrow\) & LPIPS\(\downarrow\) & \\
\midrule

Diffusion-based & LumiNet~\cite{xing2024luminet} & 16.94 & 0.614 & 0.287 & 19.94 & 0.671 & 0.274 & 17.15 & 0.781 & 0.222 & 23.82s & 13.79G\cr
Feed-forward & GPT-4o~\cite{chen2025empirical} & 14.60 & 0.418 & 0.437 & 15.13 & 0.369 & 0.415 & 12.14 & 0.479 & 0.351 & 1.36m & - \cr
Feed-forward & Ours$\dag$ & 28.35 & 0.957 & 0.031 & 29.61 & 0.947 & 0.029  & 27.48 & 0.945 & 0.051 & 0.07s & \underline{32.89M}\cr
\hline

Feed-forward & PCT-Net~\cite{guerreiro2023pct}& 22.58 & 0.912 & 0.055 & 25.26 & 0.931 & 0.035 & 25.08 & 0.921 & 0.066 & \textbf{0.03s} & \textbf{18.4M}\cr 
Diffusion-based & Objectstitch~\cite{song2023objectstitch}& 19.14 & 0.770 & 0.193 & 21.82 & 0.788 & 0.170 & 21.15 & 0.831 & 0.176& 4.54s &  5.24G \cr
Diffusion-based & ControlCom~\cite{zhang2023controlcom}& 18.85 & 0.765 & 0.209 & 19.88 & 0.771 & 0.185 & 19.75 & 0.811 & 0.189 & 4.63s &  10.94G\cr
Diffusion-based & RGB$\leftrightarrow$X~\cite{Zeng_2024} & 12.28 & 0.428 & 0.368 & 12.91 & 0.507 & 0.296 & 11.28 & 0.503 & 0.422 & 19.71s & 10.68G\cr
Diffusion-based & IC-Light~\cite{iclight}& 17.66 & 0.659 & 0.217 & 20.87 & 0.679 & 0.190 & 18.22 & 0.774 & 0.200 & 1.25s & 1.60G \cr
Feed-forward & Ours* & \underline{29.11} & \underline{0.959} & \underline{0.030} & \underline{30.00} & \underline{0.951} & \underline{0.027} & \underline{28.18} & \underline{0.945} & \underline{0.050} & 0.07s &  32.92M\cr
Feed-forward & Ours & \textbf{29.65} & \textbf{0.961} & \textbf{0.029} & \textbf{30.20} & \textbf{0.953} & \textbf{0.027} & \textbf{28.75} & \textbf{0.946} & \textbf{0.049} & \underline{0.07s} & 32.94M\cr

\bottomrule
\end{tabular}
}
\vspace{-1em}
\label{tab:single-view}
\end{table*}

\begin{figure*}[t]
\centering
\includegraphics[width=\linewidth]{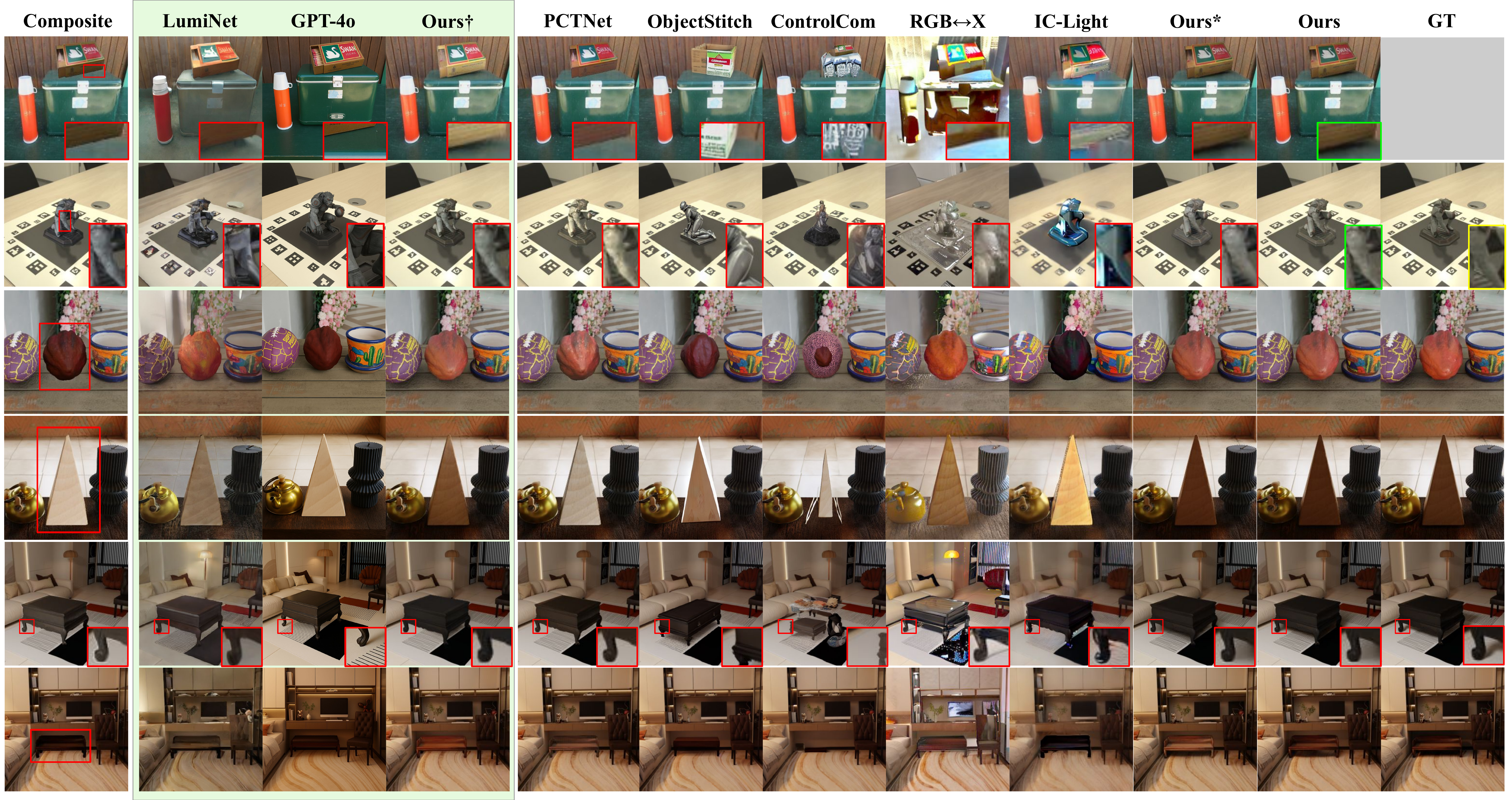}
\vspace{-1.5em}
\caption{Single-view qualitative comparison with SOTA methods~\cite{xing2024luminet, chen2025empirical, guerreiro2023pct, song2023objectstitch, zhang2023controlcom, Zeng_2024, iclight} on our proposed dataset and public datasets~\cite{zhang2023controlcom, ummenhofer2024objects}, with differences highlighted via colored patches. Compared to existing baselines, our method successfully generates illumination consistent with the background and physically plausible shadows while decoupling highlights from inserted objects, demonstrating generalization capabilities on out-of-domain datasets. The method in the green box does not incorporate background images as input, whereas the others do.}
\label{fig:singleview}
\vspace{-0.5em}
\end{figure*}

\paragraph{Baselines.}
For 2D object compositing evaluation, we conduct comprehensive comparisons with representative methods: PCTNet~\cite{guerreiro2023pct}, ObjectStitch~\cite{song2023objectstitch}, ControlCom~\cite{zhang2023controlcom}, RGB$\leftrightarrow$X~\cite{Zeng_2024}, ICLight~\cite{iclight}, LumiNet~\cite{xing2024luminet} and GPT-4o~\cite{chen2025empirical}. 
For 3D object compositing evaluation, we additionally incorporate Gaussian-based inverse rendering method such as GS-IR~\cite{liang2024gs}, GI-GS~\cite{chen2024gi}, and IRGS~\cite{gu2024irgs}, establishing a unified benchmark comparing conventional 2D pipelines with emerging 3D-aware methods built upon differentiable rendering frameworks.

\paragraph{Implementation Details.}
Our model architecture employs a unified Swin Transformer backbone with consistent configurations for both 2D and 3D object compositing tasks. The network processes 256×256 resolution inputs with an embedding dimension of 96, structured with 3 cascaded transformer blocks. Each block contains 6 successive Swin Transformer layers with 6 parallel attention heads. 

We train the 2D object compositing model for 1M iterations with batch size 128 using AdamW (base lr=2e-3, weight decay=0.05, momentum parameters $\beta_1$=0.9, $\beta_2$=0.95), 10k iteration linear warmup followed by cosine decay to 1e-6, FP16 mixed precision, gradient clipping at 10.0, and an EMA of 0.99. 
For 3D object compositing model, we reduce the learning rate to 1e-3 and batch size to 32, training for 100k iterations. 
For trainig time, We train the 2D model for ~15 days, and the 3D model for ~3 days with 16 NVIDIA A100 (80 GB) GPUs.

\subsection{Performance Analysis}

\begin{table*}[t!]
\renewcommand{\arraystretch}{1.15}
\setlength{\tabcolsep}{3pt}
\centering
\caption{Multi-view quantitative performance on our purposed dataset and real captured scenes.  We report visual quality metrics, inference time (Gaussian training time \textit{\#Train}), highlighting the \textbf{best} and \underline{second-best} in each category. Our* and Our$\dag$ denote our method without depth input and without both depth and background input, respectively.}
\resizebox{1\linewidth}{!}{
\begin{tabular}{c|c|ccc|ccc|ccc|c}
\toprule
\multicolumn{2}{c|}{Dataset} &  \multicolumn{3}{c|}{Simple Synthetic Scene} & \multicolumn{3}{c|}{Complex Synthetic Scene} & \multicolumn{3}{c|}{Real Captured Scene} & \multirow{2}{*}{Time\(\downarrow\) (\#Train)} \\ 
 Paradigm & Method & PSNR\(\uparrow\) & SSIM\(\uparrow\) & LPIPS\(\downarrow\) & PSNR\(\uparrow\) & SSIM\(\uparrow\) & LPIPS\(\downarrow\) & PSNR\(\uparrow\) & SSIM\(\uparrow\) & LPIPS\(\downarrow\) \\
\midrule

Diffusion-based & LumiNet~\cite{xing2024luminet} & 17.15 & 0.573 & 0.304 & 18.45 & 0.663 & 0.222 & 20.05 & 0.770 & 0.198 & 6.31m (-) \cr
Feed-forward & GPT-4o~\cite{chen2025empirical} & 14.57 & 0.375 & 0.445 & 15.11 & 0.366& 0.411 & 14.34 & .473 & 0.406 & 21.40m (-) \cr
Feed-forward & Ours$\dag$ & 28.96 & 0.955 & 0.033 & 29.32 & 0.946 & 0.029 & 25.88 & 0.925 & 0.041 & 1.07s (1.08m)\cr
\hline
Feed-forward & PCT-Net~\cite{guerreiro2023pct} & 22.97 & 0.908 & 0.057 & 25.19 & 0.927 & 0.035 &  23.39 & 0.824 & 0.103 & \textbf{0.47s} (-) \cr
Diffusion-based & Objectstitch~\cite{song2023objectstitch}& 19.12 & 0.726 & 0.217 & 21.84 & 0.792 & 0.163 & 18.43 & 0.785 & 0.193 & 1.21m (-) \cr
Diffusion-based & ControlCom~\cite{zhang2023controlcom}& 18.95 & 0.722 & 0.231 & 19.60 & 0.773 & 0.181 & 18.54 & 0.778 & 0.199 &  1.23m (-) \cr
Diffusion-based & RGBX~\cite{Zeng_2024} & 12.71 & 0.417 & 0.360 & 12.69 & 0.504 & 0.304 & 13.68 & 0.594 & 0.312 & 5.26m (-) \cr
Diffusion-based & IC-Light~\cite{iclight} & 17.94 & 0.596 & 0.242 & 20.60 & 0.689 & 0.183 & 20.23 & 0.718 & 0.233 & 19.36s (-) \cr
Inverse Rendering & GS-IR~\cite{liang2024gs} & 15.56 & 0.742 & 0.134 & 16.81 & 0.664 & 0.249 & 15.92 & 0.699 & 0.265 & 17.62s (57.14m) \cr
Inverse Rendering &GI-GS~\cite{chen2024gi} & 18.97 & 0.808 & 0.126 & 16.56 & 0.674 & 0.310 
& 16.07 & 0.716 & 0.234 & 16.69s (1.43h)\cr
Inverse Rendering & IRGS~\cite{gu2024irgs} & 17.79 & 0.688 & 0.237 & 21.04 & 0.702 & 0.291 & 20.19 & 0.744 & 0.215 & 9.72m (3.02h) \cr
Feed-forward & Ours* & \underline{29.73} & \underline{0.958} & \underline{0.031} & \underline{29.51} & \underline{0.949} & \underline{0.028} & \underline{26.10} & \underline{0.926} & \underline{0.041} & \underline{1.07s (1.08m)}\cr
Feed-forward & Ours & \textbf{30.29} & \textbf{0.960} & \textbf{0.030} & \textbf{30.13} & \textbf{0.952} & \textbf{0.027} & \textbf{26.39} & \textbf{0.927} & \textbf{0.040} & 1.08s (1.08m) \cr

\bottomrule
\end{tabular}
}
\vspace{-1em}
\label{tab:multi-view}
\end{table*}

\begin{figure*}[t]
\centering
\includegraphics[width=\linewidth]{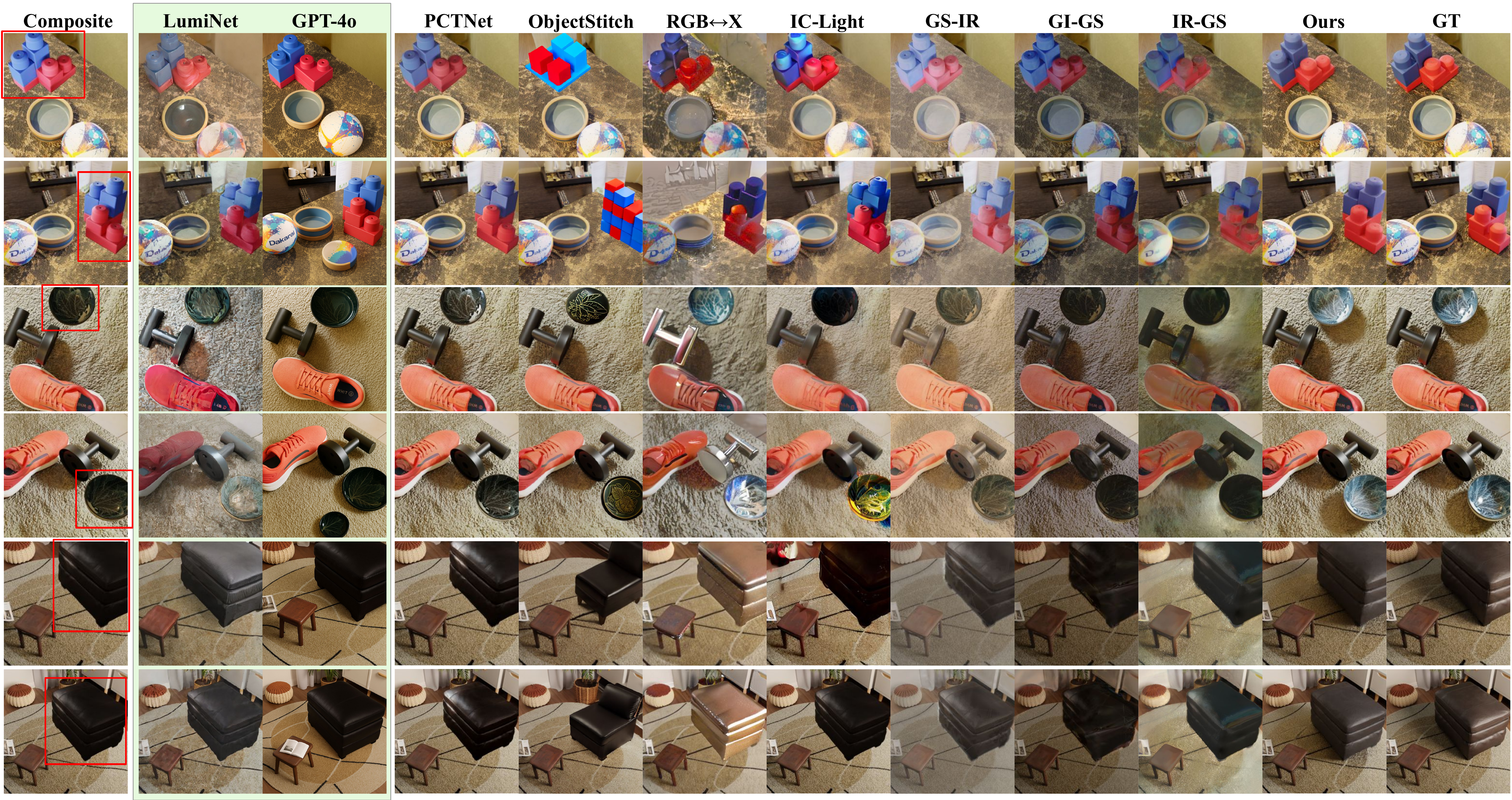}
\vspace{-1.5em}
\caption{Multi-view qualitative comparison with SOTA methods~\cite{xing2024luminet, chen2025empirical, guerreiro2023pct, song2023objectstitch, zhang2023controlcom, Zeng_2024, iclight, liang2024gs, chen2024gi, gu2024irgs} on our proposed dataset and real captured scenes, with differences highlighted via colored patches. Our method synthesizes plausible illumination and shadows while ensuring multi-view consistency. The method in the green box does not incorporate background images as input, whereas the others do.}
\vspace{-0.5em}
\label{fig:multiview}
\end{figure*}

\begin{figure}[t]
\centering
\includegraphics[width=\linewidth]{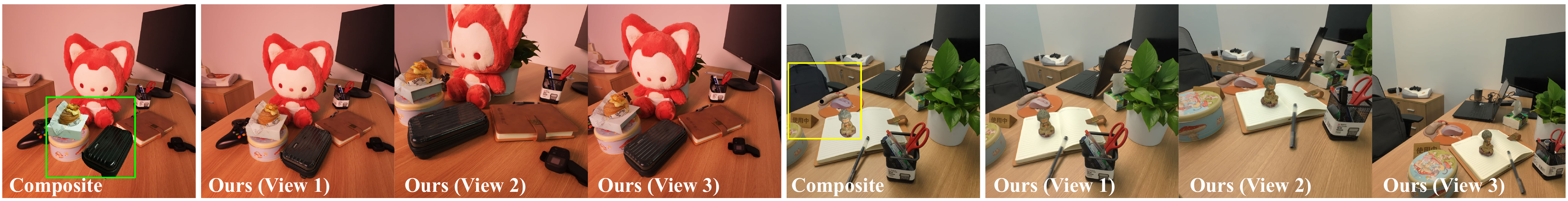}
\caption{
We evaluate our method on real-world scenes captured under varying illumination with six cameras arranged in a circular array. 
On the left, we insert a cake and a black box; on the right, we insert a toy, a mouse, and a backpack. 
MV-CoLight consistently harmonizes object colors, produces physically plausible lighting interactions, and accurately casts shadows across all viewpoints.
}
\vspace{-1em}
\label{fig:realcap}
\end{figure}

Below we show our method delivers physically plausible lighting and shadows for inserted objects, out-performing both 2D harmonization~\cite{xing2024luminet, chen2025empirical, guerreiro2023pct, song2023objectstitch, zhang2023controlcom, Zeng_2024, iclight} and Gaussian-based inverse rendering~\cite{liang2024gs, chen2024gi, gu2024irgs} baselines.
The approach also generalizes from synthetic training to challenging real-world captures, maintaining photorealism under diverse lighting and materials.

\paragraph{Single-view Harmonized Result.}

Current image harmonization methods exhibit notable limitations 
when compositing new objects into a scene, as illustrated in Fig.~\ref{fig:singleview}.
For example, PCT-Net\cite{guerreiro2023pct} enforces only color consistency and omits realistic highlights and cast shadows, while RGB-X~\cite{Zeng_2024} material estimation yields inaccurate albedo maps that blur illumination and misalign geometry during neural relighting.
Diffusion-based frameworks such as ObjectStitch~\cite{song2023objectstitch} and ControlCom~\cite{zhang2023controlcom} produce visually compelling composites but often distort object shape and texture in the generative process.
ICLight~\cite{iclight}'s illumination estimator lacks robustness in cross-domain scenarios, resulting in pronounced appearance artifacts under complex real-world lighting, while the light transport module of LumiNet~\cite{xing2024luminet} generates non-physical highlight patterns and jagged shadow boundaries.
Even advanced multimodal systems like GPT-4o~\cite{chen2025empirical}, which improve local lighting coherence, introduce unintended global modifications that undermine overall scene integrity which is particularly hard to be strictly enforced via prompting.

\paragraph{Multi-view Harmonized Result.}

Multi-view object compositing compounds the inconsistencies of 2D harmonization methods, resulting in visible color shifts and misaligned shadows across viewpoints. 
Gaussian-based inverse rendering techniques attempt to remedy this by enriching each primitive with material attributes, such as estimated albedo and normals, and estimating an environment map from the background Gaussians to relight the composite. 
However, their reliance on imperfect decoupling causes specular highlights and shadowed regions from the original images to be treated as textures. 
As shown in Fig.~\ref{fig:multiview}, the result is a conflated relighting effect that blurs the distinction between intrinsic material properties and new environmental illumination, failing to achieve true multi-view coherence.
Unlike environment mapping-based methods, our approach directly learns illumination and shadow priors from the multi‐view composite scene and transfers them to the inserted object, guaranteeing seamless, view‐consistent lighting. 
By encoding these learned visual cues into Gaussian feature representations and propagating them through our transformer-based 2D-3D pipeline,  we maintain spatial coherence and realistic shadowing without explicit environment map estimation. Extensive evaluations on public benchmarks~\cite{zhang2023controlcom, ummenhofer2024objects} and our own dataset demonstrate that our method outperforms both 2D harmonization and Gaussian‐based inverse‐rendering baselines in quantitative metrics and visual quality.

\paragraph{Real Scene Harmonized Result.}
We further assess our method on diverse real-world multi-view captures that diverge markedly from our synthetic training data in terms of lighting complexity and material detail, as shown in Fig.~\ref{fig:realcap}. These scenes feature unpredictable illumination conditions and intricate textures that typically confound traditional harmonization and inverse-rendering pipelines. Nevertheless, our approach consistently produces photorealistic composites, accurately estimating lighting and casting coherent shadows across all viewpoints. This robust performance under uncontrolled, real-world conditions highlights the generalization robustness of our learning-based illumination priors.

\subsection{Efficiency and Extensibility}

\paragraph{Inference Time Comparison.} 
For 2D object compositing, our feedforward architecture enables inference times as short as about 0.07 seconds, significantly outperforming diffusion-based methods like LumiNet~\cite{xing2024luminet}, which require more than 20 seconds to complete the multi-step denoising process (default 50 steps). 
While GPT-4o~\cite{chen2025empirical} remains closed-source, we utilize its web interface to generate results, incurring a latency of several minutes. For 3D object compositing, Gaussian-based inverse rendering methods necessitate both environmental map extraction and material attribute optimization atop pre-trained Gaussians. 
In contrast, our method achieves harmonized Gaussian color attributes with about 1 second of inference time after a few minutes of scene-specific Gaussian representation learning, demonstrating superior efficiency. Notably, our framework eliminates the need for Gaussian retraining when repositioning inserted objects. This efficiency makes our framework especially well suited for real-time AR and embodied-intelligence applications.

\begin{figure*}[t]
\centering
\includegraphics[width=\linewidth]{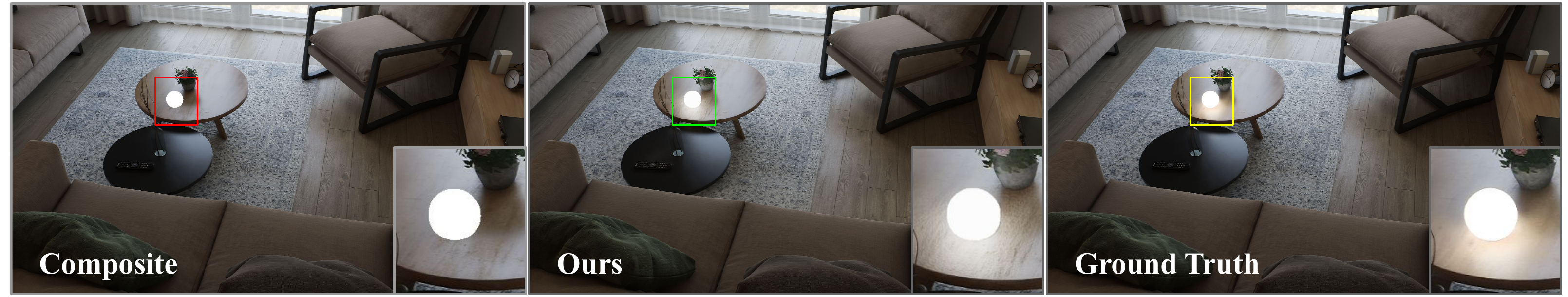}
\vspace{-1em}
\caption{Visual results of inserting luminous objects. Our method successfully simulates the illumination effects of luminous spheres within the scene environment.}
\label{fig:lighting}
\vspace{-0.5em}
\end{figure*}

\paragraph{Extentions on Various Illumination Priors.}
By design, our unified compositing pipeline can also accommodate other illumination priors - whether HDR environment maps, learned light distributions, or discrete emitters, with similar training and inference pipeline. 
This challenging capability has been largely neglected by prior works.
As a demonstrative extension in Fig.~\ref{fig:lighting}, we focus here on inserting new light sources to dynamically relight the scene. To support emissive-object compositing with true multi-view consistency, our method estimates light propagation through the existing 3D Gaussian geometry, capturing how point or area emitters illuminate surrounding surfaces, and computes secondary shadows and interreflections to generate physically plausible shading on all objects. This extension underscores the flexibility of our approach and its applicability to a wide range of illumination scenarios.

\subsection{Ablation Study}
We perform a series of ablations to isolate the factors driving our 2D and 3D compositing pipelines, the results are shown in Tab.~\ref{tab:abs}. First, we vary the number of swin transformer blocks, layers and the feature‐embedding dimension. We find that embedding size has a greater impact on final performance than the depth of the transformer stack, with a modest drop in harmonization quality when reducing the block number or layer number. 
Next, we examine the 2D compositing inputs, removing the background image prevented reliable object placement, while omitting the depth map eliminated essential geometric priors.
We find that both scenarios degrade the model’s ability to learn illumination conditions. 
When we feed these 2D features into the 3D network, excluding them entirely still allowed plausible color matching but produced unrealistic shadows and highlights, underscoring the importance of 2D illumination cues. Introducing the Hilbert curve reordering further accelerate training convergence and improve visual quality by preserving Gaussian color locality in 2D Hilbert space.

\begin{table*}[t!]
\renewcommand{\arraystretch}{1.15}
\setlength{\tabcolsep}{4pt}
\centering
\caption{Ablation study conducted on the simplified synthetic scenes within our proposed dataset. We report visual quality metrics, inference time and memory storage.}
\resizebox{1\linewidth}{!}{
\begin{tabular}{l|ccccc|ccccc}
\toprule
Model & \multicolumn{5}{c|}{2D Object Compositing Model} & \multicolumn{5}{c}{3D Object Compositing Model} \\ 
Method & PSNR\(\uparrow\) & SSIM\(\uparrow\) & LPIPS\(\downarrow\) & Time\(\downarrow\) & Memory\(\downarrow\) & PSNR\(\uparrow\) & SSIM\(\uparrow\) & LPIPS\(\downarrow\) & Time\(\downarrow\) & Memory\(\downarrow\) \\
\midrule
baseline & 29.65 & 0.961 & 0.029 & 0.07s & 32.94M & 30.29 & 0.960 & 0.030 & 1.08s & 35.01M\cr
\hline
transformer block (2) & 28.34 & 0.955 & 0.032 & 0.06s & 23.27M & 28.70 & 0.953 & 0.035 & 0.97s & 25.34M \cr
transformer layer (4) & 28.39 & 0.956 & 0.032 & 0.06s & 23.38M & 28.83 & 0.957 & 0.034 & 0.98s & 25.45M \cr
embedding dim (60) & 27.68 & 0.951 & 0.035 & 0.05s & 14.06M & 28.11 & 0.949 & 0.039 & 0.81s & 14.17M \cr
\hline
w/o Hilbert transform& - & - & - & - & - & 28.99 & 0.951 & 0.036 & 1.08s & 35.01M \cr\
w/o 2D OC model & - & - & - & - & - & 25.83 & 0.913 & 0.051 & 1.08s & 35.01M \cr
w/o depth input & 29.11 & 0.959 & 0.030 & 0.07s & 32.92M & 29.73 & 0.958 & 0.031 & 1.07s & 34.99M \cr
w/o background input & 28.81 & 0.958 & 0.030 & 0.07s & 32.91M & 29.44 & 0.956 & 0.032 & 1.07s & 34.98M\cr

\bottomrule
\end{tabular}
}
\vspace{-1em}
\label{tab:abs}
\end{table*}

\section{Conclusion}
\label{sec:conclusion}
In this paper, we introduce MV-CoLight, a two-stage framework that seamlessly combines a 2D feed-forward harmonization network with a 3D Gaussian-based compositing model to deliver efficient, view-consistent object insertion. In the first stage, our 2D network rapidly learns per-view color and illumination alignment; in the second, the 3D Gaussian fields enforce geometric and lighting coherence across viewpoints, producing realistic shadows and reflections with minimal runtime overhead. Extensive experiments on both synthetic and real-world benchmarks show that MV-CoLight outperforms state-of-the-art 2D and 3D baselines in visual fidelity and consistency. 
To drive further progress, we also introduce a new large-scale multi-view compositing dataset with photorealistic accurate annotations.
Finally, we demonstrate that our pipeline naturally generalizes to additional lighting effects, underscoring its versatility for broader applications. 

\bibliographystyle{plain}
\bibliography{main}

\newpage

\appendix
\section{Supplementary Material}
In the supplementary material, we first present a brief overview of the core concepts underlying Gaussian splatting~\cite{kerbl3Dgaussians} and the Hilbert curve~\cite{hilbertcurve} in Sec.~\ref{supp:preliminaries}. Subsequently, Sec.~\ref{supp:dataset}  elaborates in detail on the dataset construction process and preprocessing procedures.
Furthermore, the details of our object compositing models are elaborated in Sec.~\ref{supp:detailed}, along with the reason why we select this specific model architecture. In Sec.~\ref{supp:us} and Sec.~\ref{supp:visual}, we present a user study and additional qualitative results demonstrating objectively and subjectively  outcomes in both single-view and multi-view object compositing across multiple public datasets and our proposed dataset. We further showcase multi-view visualization for illuminative object insertion. Sec.~\ref{supp:trend} illustrates the effectiveness of different training stages of the model and analyzes the training dynamics of the 2D compositing model. Moreover, for unposed inputs, we present multi-view object insertion results based on camera poses and depth maps provided by VGGT~\cite{wang2025vggt}, as shown in Sec.~\ref{supp:vggt}. Finally in Sec.~\ref{supp:limitation}, we discuss  the limitations of our approach and outline potential directions for future research.

\subsection{Preliminaries}
\label{supp:preliminaries}

\paragraph{Gaussian Splatting} 
\label{pre: gaussain}
3D Gaussian splatting~\cite{kerbl3Dgaussians} models 3D scenes using anisotropic Gaussian primitives, employing a projection-based rasterization process to generate photorealistic renderings. Each primitive is mathematically represented by a multivariate Gaussian distribution parameterized as:  

\begin{equation}
    G(\mathbf{x}) = \exp\left(-\frac{1}{2}(\mathbf{x}-\mu)^\top \Sigma^{-1}(\mathbf{x}-\mu)\right)
\end{equation}  

where \(\mu \in \mathbb{R}^3\) specifies the spatial centroid and \(\Sigma \in \mathbb{R}^{3\times 3}\) denotes the covariance matrix. For geometric interpretability, the covariance matrix is factorized into rotation and scaling components through the decomposition \(\Sigma = RSS^\top R^\top\), where \(R \in \text{SO}(3)\) represents the rotation matrix and \(S = \text{diag}(s_x, s_y, s_z)\) encodes axis-aligned scaling factors.  

During differentiable rendering, each Gaussian primitive is augmented with additional attributes, an opacity coefficient \(\sigma \in [0,1]\) controlling light transmittance and a spherical harmonics \(F \in \mathbb{R}^C\)  enable view-dependent color estimation $c \in \mathbb{R}^3$ through directional decoding. 
The rendering process employs a tile-based rasterization pipeline that first performs efficient depth sorting of Gaussians in camera-facing order, followed by perspective projection to transform 3D Gaussian distributions into 2D image-plane counterparts \( G'(\mathbf{x}') \), and finally executes per-pixel \(\alpha\)-compositing through the rendering equation:

\begin{equation}
    C(\mathbf{x}') = \sum_{i \in \mathcal{N}} T_i c_i  \alpha_i, \quad \alpha_i = \sigma_i G'_i(\mathbf{x}')  
\end{equation}  

where $x^{\prime}$ is the queried pixel, $\mathcal{N}$ represents depth-sorted sequence of 2D Gaussians associated with $x^{\prime}$ and $T$ denotes the cumulative transmittance term as $\prod_{j=1}^{i-1}\left(1-\alpha_j\right)$.

\paragraph{Hilbert Curve}
the Hilbert curve\cite{hilbertcurve} is a space-filling curve that maps multidimensional data to a one-dimensional sequence while preserving locality. The Hilbert curve recursively partitions the space, ensuring that points close in the multidimensional domain remain near each other in the one-dimensional ordering. This locality preservation is critical for maintaining the spatial coherence of Gaussian primitives during their projection onto a 2D image space.

\begin{figure*}[t!]
\centering
\includegraphics[width=\linewidth]{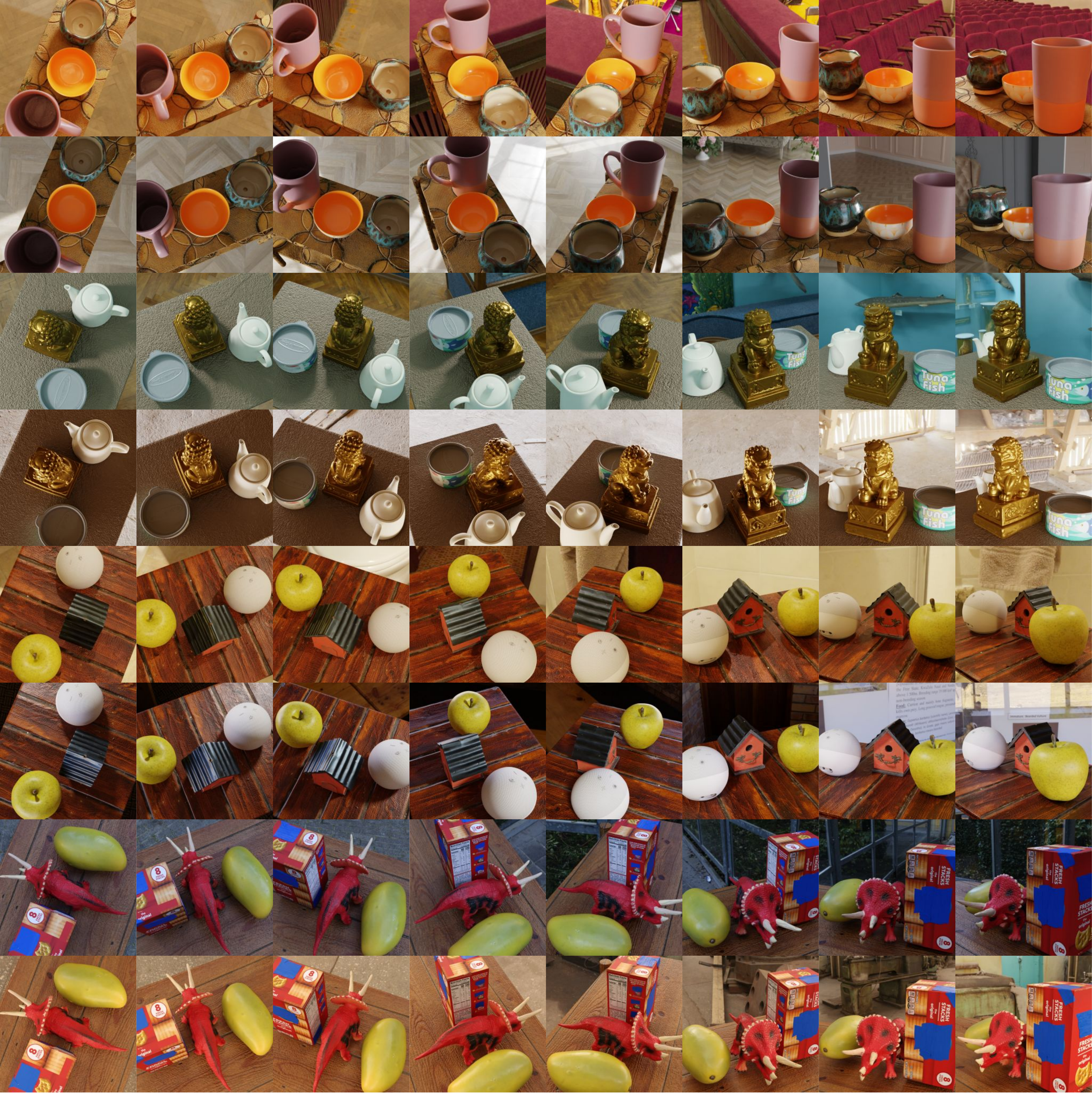}
\caption{
Visualization of the DTC-MultiLight dataset. We showcase rendered results of diverse scenes created using objects from the DTC dataset within the Blender engine, highlighting multi-view perspectives and varying lighting conditions.}
\label{fig:dataset}
\end{figure*}

\begin{figure*}[t!]
\centering
\includegraphics[width=\linewidth]{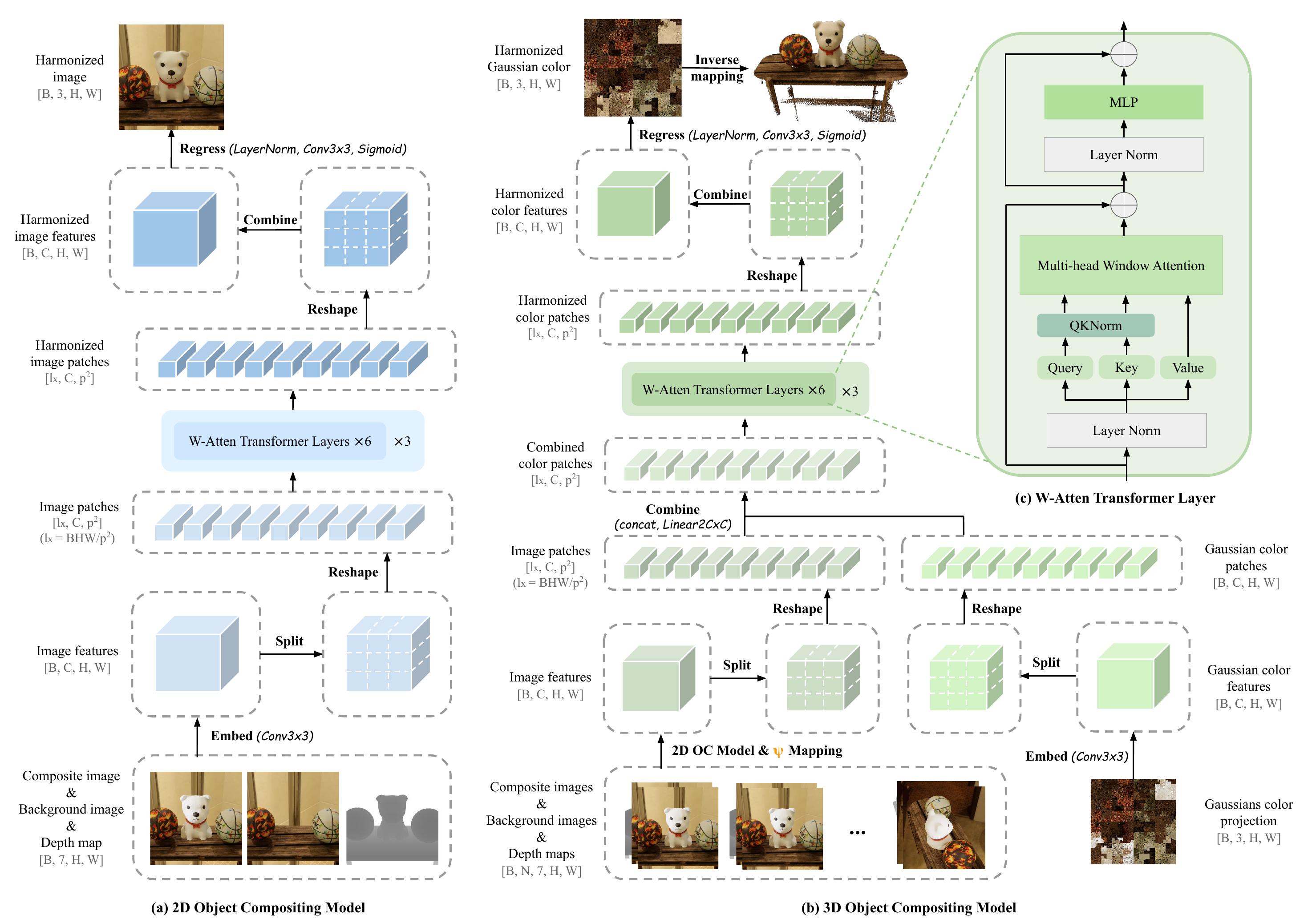}
\caption{Detailed model architecture of our compositing models. We introduce two compositing models: (a) The 2D object compositing model encodes input data into shallow feature space, which is then partitioned into $p \times p$ size patches. These features are decoded through a Swin Transformer-based decoder to produce harmonized output. (b) The 3D object compositing model encodes the output features from the 2D model and inharmonious Gaussian colors. Through a similar process, it generates harmonized Gaussian colors and reprojects them into the Gaussian model. Both models employ pure window-attention layers as shown in (c).}
\label{fig:detail}
\end{figure*}

\begin{figure*}[t!]
\centering
\includegraphics[width=\linewidth]{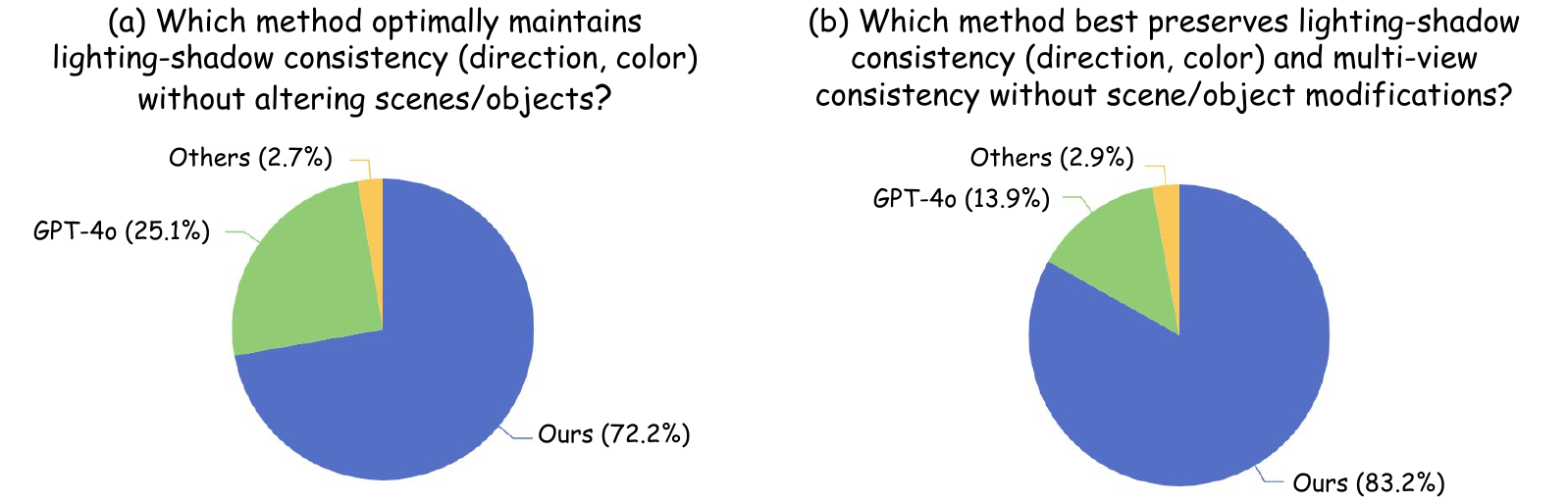}
\caption{User Study Results. We respectively compare our method with baseline methods for both single-view object compositing and multi-view object compositing to quantify the realism in lighting and shadow generation. Results demonstrate that our method outperforms baseline methods.}
\label{fig:user_study}
\end{figure*}

\subsection{Dataset Curation}
\label{supp:dataset}
We present a large-scale synthetic dataset, named \textbf{DTC-MultiLight},  specifically designed for consistent object compositing in 3D scenes, comprising about 480,000 procedurally generated scenes with systematically varied scene components and illumination conditions, as shown in Fig.~\ref{fig:dataset}. This comprehensive dataset, created using Blender's rendering engine, serves as a robust benchmark for both training and evaluating object compositing models across diverse object placements and illumination settings.

When constructing object repositories for scene composition, we select the Digital Twin Catalog (DTC) dataset~\cite{dong2025digital} over the widely adopted Objaverse~\cite{deitke2023objaverse} due to its superior scanning accuracy, which enables more physically accurate simulations of lighting interactions and shadow casting. From this dataset, we curate four distinct tables serving as placement surfaces, complemented by 1,752 diverse household objects. 
However, we observe that the original table surface materials exhibit insufficient albedo values for clear shadow visualization. To address this visual limitation, we implement 69 distinct material properties from Poly Haven (primarily wood and concrete textures). This material diversification strategy serves to enhance dataset variability while improving the model's generalizability to real-world surface reflectance conditions.
However, we observe that the original table surface materials exhibit insufficient albedo values for clear shadow visualization. To address this visual limitation, we implement 69 different physically-based material presets from Poly Haven ( encompassing wood and concrete textures). This material diversification strategy serves to enhance dataset variability while improving the model's generalizability to real-world surface reflectance conditions.
To establish photorealistic illumination conditions and enhance contextual background elements, we integrate 207 indoor high-dynamic-range (HDR) environment maps sourced from Poly Haven\footnote{\url{https://polyhaven.com/}}. Recognizing the inherent limitations of environment maps in generating sharp shadow boundaries, we strategically augment scenes with supplementary light sources to achieve enhanced lighting variation and directional shadow effects.

During the Blender rendering process, we randomly select three distinct objects, placing them on the table in a random arrangement. Under the illumination of a random environment map and additional light sources, we render 16 images from specific perspectives centered on the objects, striving to cover the scene comprehensively, as shown in Fig.~\ref{fig:dataset}.
Furthermore, we intentionally assign emissive material properties to the inserted objects in a subset of our dataset, enabling these luminous entities to physically influence the ambient illumination conditions of background scenes, augmenting the diversity and challenging of our dataset.
Ultimately, we develop a comprehensive multi-view object compositing dataset featuring varied illumination scenarios, heterogeneous object categories, and diverse background materials. 
This versatile dataset demonstrates broad applicability across multiple computer vision domains, such as multi-view object compositing, scene relighting and scene generation.

To simulate object insertion, we first randomly select two sets of multi-view images captured under different lighting conditions for the same scene. We then use mask maps to separate each image into foreground and background, where the foreground contains the same object across both sets. Finally, we merge the foreground and background from the same viewpoint but under different lighting conditions to simulate object insertion. The mask maps are generated by rendering the scenes in Blender, where the foreground object and background are assigned distinct colors to facilitate their separation.

\subsection{Detailed Model Architecture}
\label{supp:detailed}
 We have provided a detailed model architecture figure, as shown in Fig.~\ref{fig:detail}. 
 For the 2D object compositing model, we built upon the original Swin-Transformer architecture by adding CNN-based feature extraction and output modules to align the input/output dimensionalities. Additionally, we incorporated residual modules in the feature space to stabilize the model's learning of scene harmonization. Subsequently, we utilized multiple Swin Transformer blocks to extract color attributes, illumination properties, and spatial shadow relationships from shallow features.
For the 3D object compositing model, we projected the output features of the 2D model into a Gaussian color-aligned 2D space, fused inconsistent Gaussian color representations during the encoding phase, and employed Swin Transformer blocks with identical structures to address 3D scene harmonization. Finally, the Gaussian color outputs from the model were back-projected into the Gaussian space.

By choosing the more efficient Swin Transformer over the original ViT, we accelerated both training and inference speeds while increasing the upper limits for input quantity and resolution. The local attention mechanism further directs focused attention to detailed highlights and shadow generation. As shown in Figure 8, our method even captures fine-grained texture shadows on inserted objects.

\subsection{User Study}
\label{supp:us}
Our method surpasses all baseline methods under objective criteria, as demonstrated in Fig.~\ref{fig:user_study}.

\subsection{Additional Visual Results}
\label{supp:visual}
\begin{figure*}[t!]
\centering
\includegraphics[width=\linewidth]{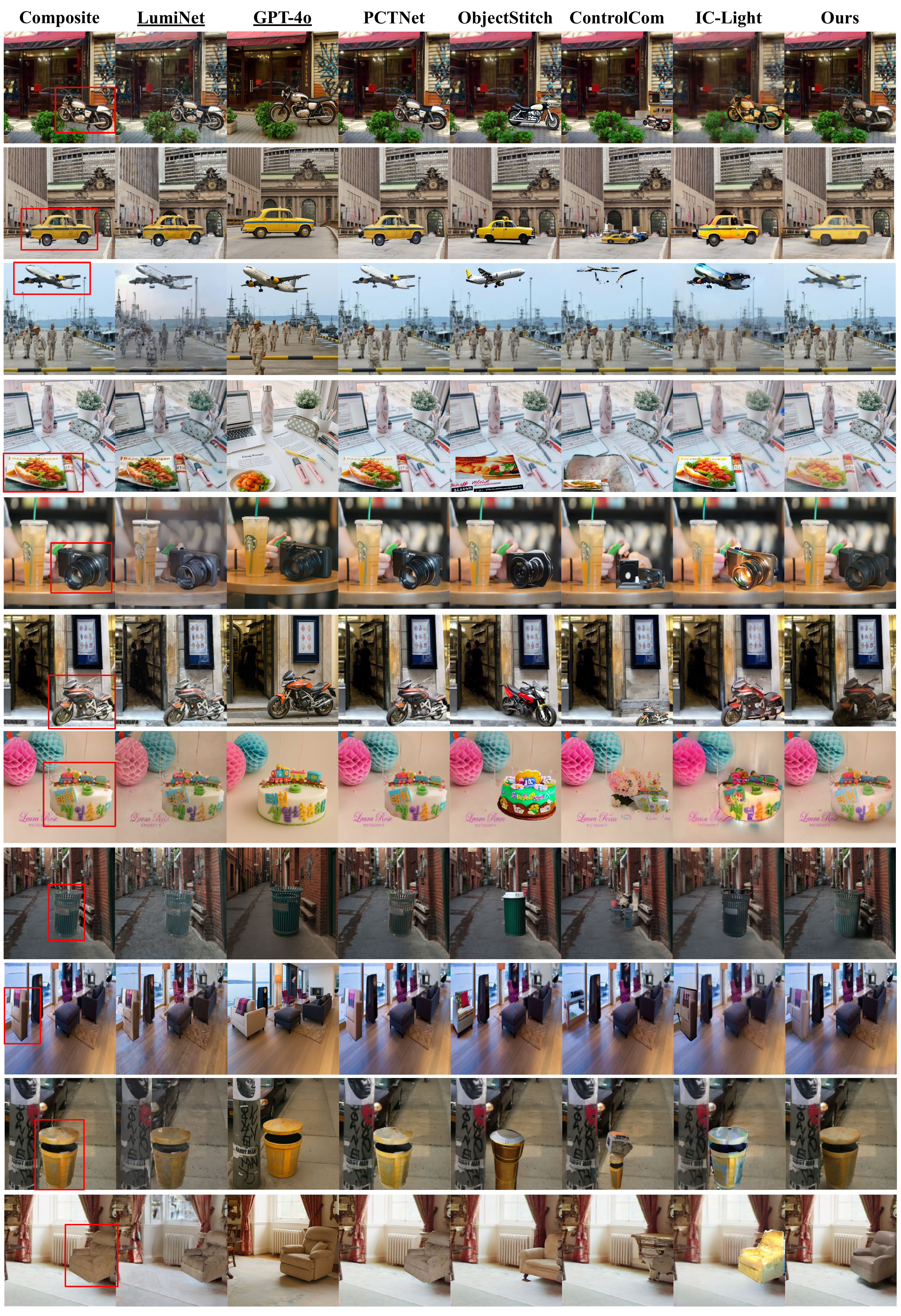}
\caption{
Single-view qualitative comparison on the Foscom dataset~\cite{zhang2023controlcom}. Our method not only performs color harmonization but also generates realistic highlights and shadows, providing more visually convincing results compared to baseline methods such as PCTNet~\cite{guerreiro2023pct}, which focus solely on color harmonization. The \underline{methods} do not require background image input, while others include.}
\label{fig:foscom_supp}
\end{figure*}

\begin{figure*}[htbp]
\centering
\includegraphics[width=\linewidth]{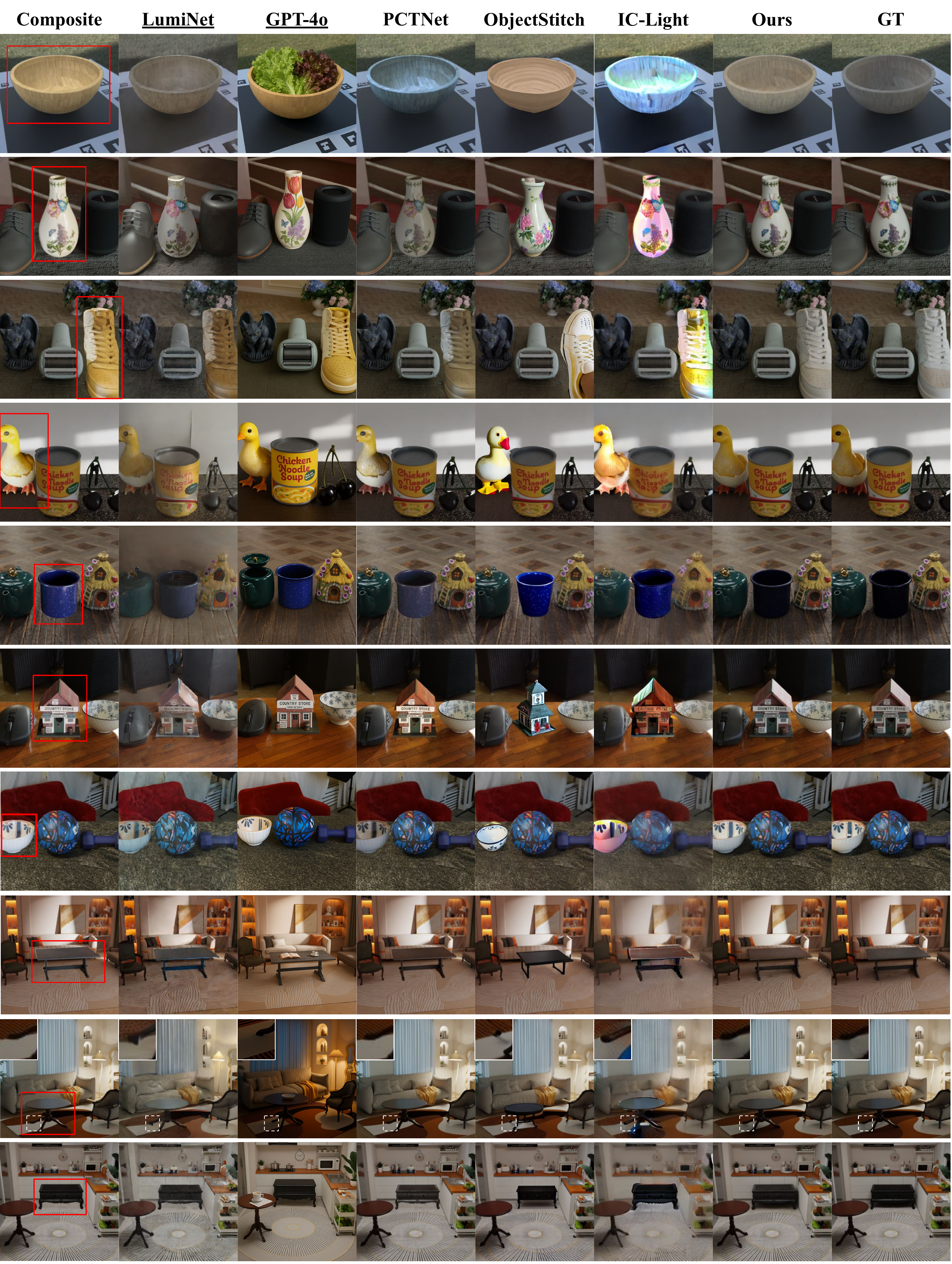}
\caption{
Single-view qualitative comparison on the Object with Lighting dataset~\cite{ummenhofer2024objects} and our proposed dataset. 
Our approach achieves implicit lighting disentanglement for inserted objects, synthesizing spatially consistent illumination and shadows that adaptively align with background lighting conditions. Unlike baseline methods, our framework produces photorealistic lighting effects surpassing existing approaches in both object-centric simple scenes and indoor complex environments, while strictly preserving the original scene geometry, scale, and object positioning.
The \underline{methods} do not require background image input, while others include.
}
\label{fig:single_supp}
\end{figure*}

\begin{figure*}[t!]
\centering
\includegraphics[width=\linewidth]{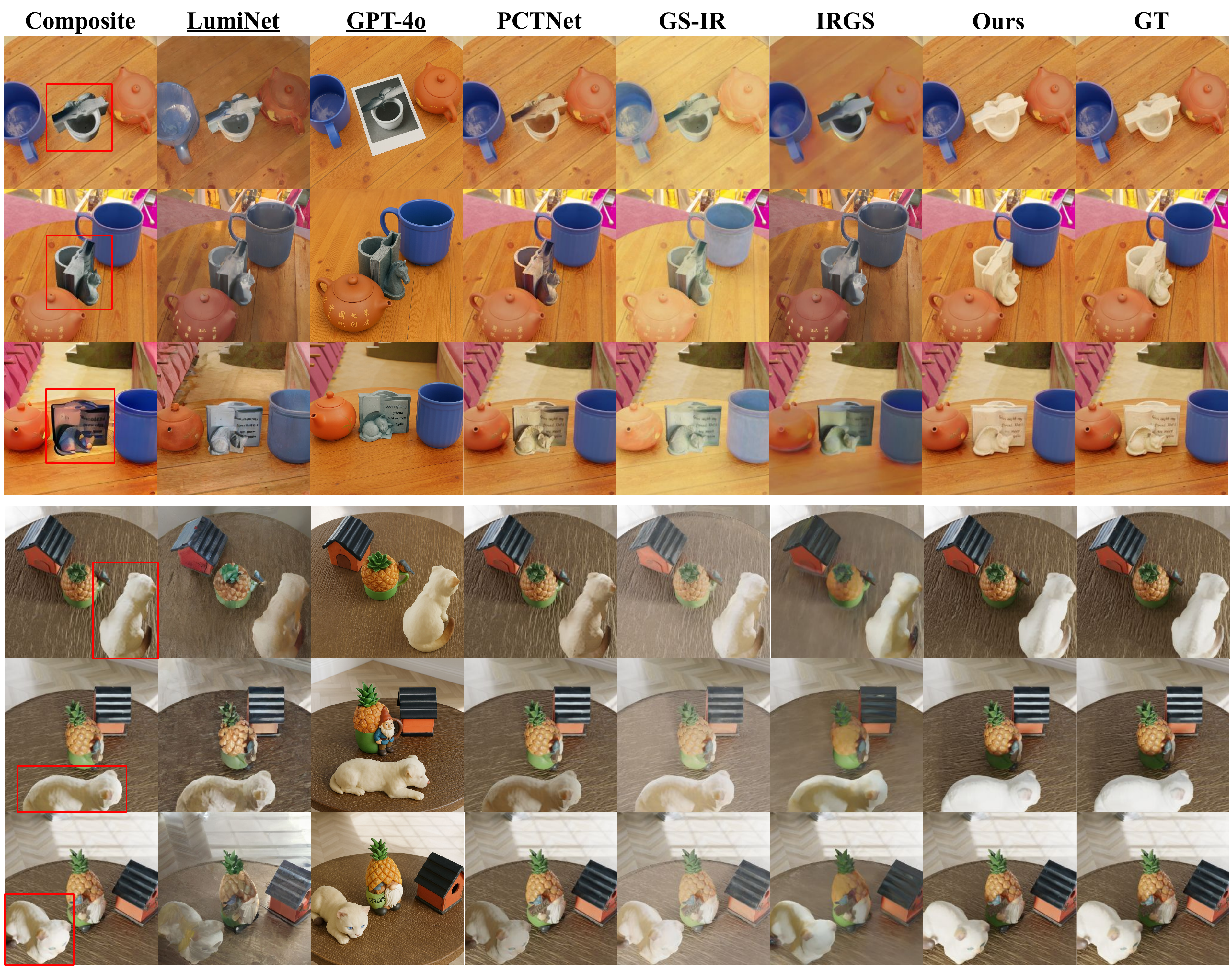}
\caption{Multi-view qualitative comparison on our rendered scenes. 
In the first case, we remove existing shadows on the inserted object. In the second case, while eliminating shadows, we generate new shadows on the left side of the kitten and corresponding desktop based on the top-right light source in the scene. In addition, our method achieves multi-view consistency, whereas  those methods that focus on image-level tasks exhibit color and shape discrepancies. Gaussian-based inverse rendering, constrained by environment map inputs, produces overly bright or dark visual artifacts.
The \underline{methods} do not require background image input, while others include.
}
\label{fig:multi_supp}
\end{figure*}

\begin{figure*}[t!]
\centering
\includegraphics[width=\linewidth]{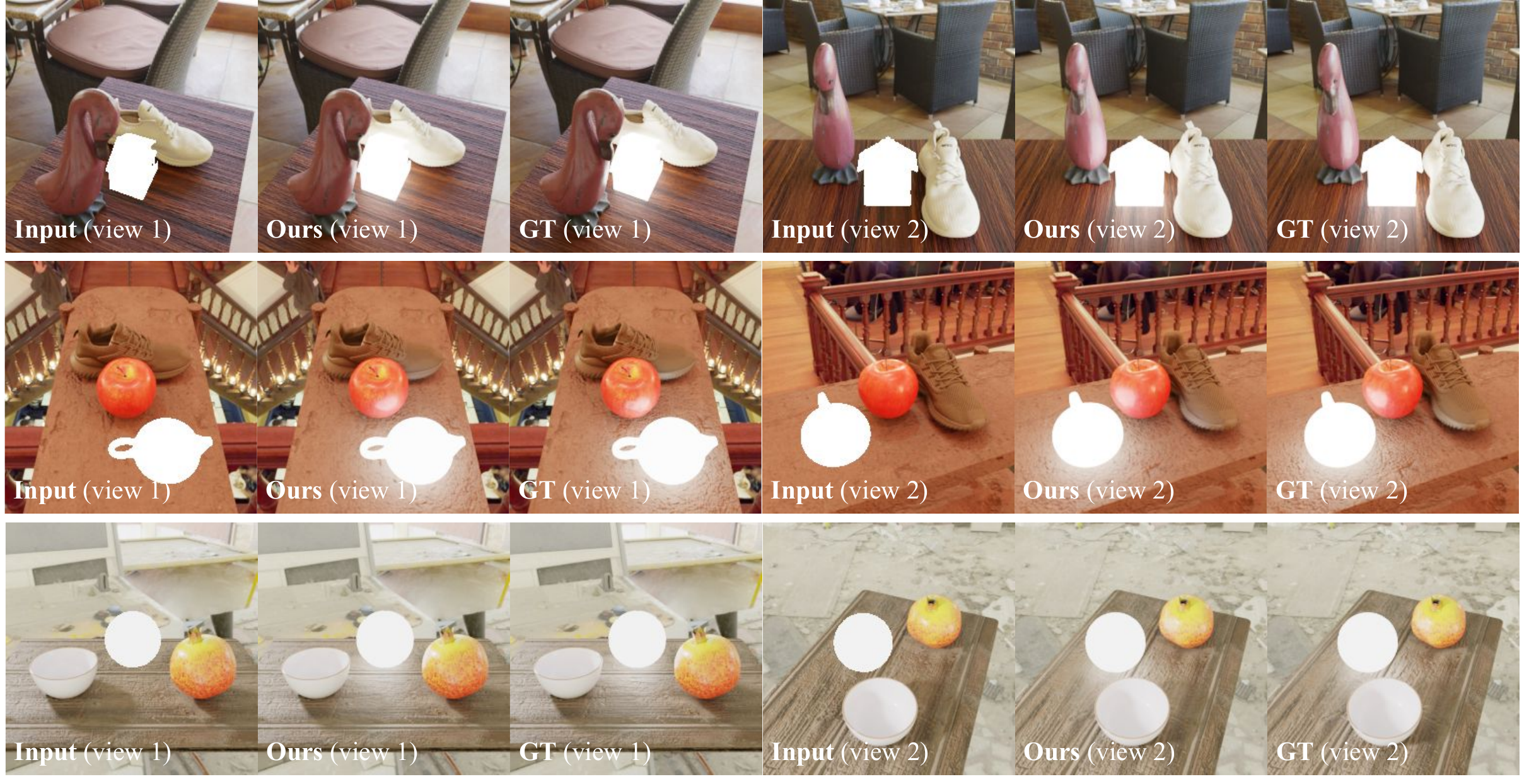}
\caption{Multi-view visualization after light source insertion. Our method meticulously simulates the emission effects of inserted light sources, their illumination on surrounding objects, and shadows.}
\label{fig:light_supp}
\end{figure*}

We provide additional visualizations comprising single-view object compositing (Fig.~\ref{fig:foscom_supp},~\ref{fig:single_supp}), multi-view object compositing (Fig.~\ref{fig:multi_supp}), and multi-view light insertion (Fig.~\ref{fig:light_supp}). Comprehensive qualitative comparisons substantiate the superiority and robustness of our approach.

\clearpage

\begin{figure*}[t!]
\centering
\includegraphics[width=\linewidth]{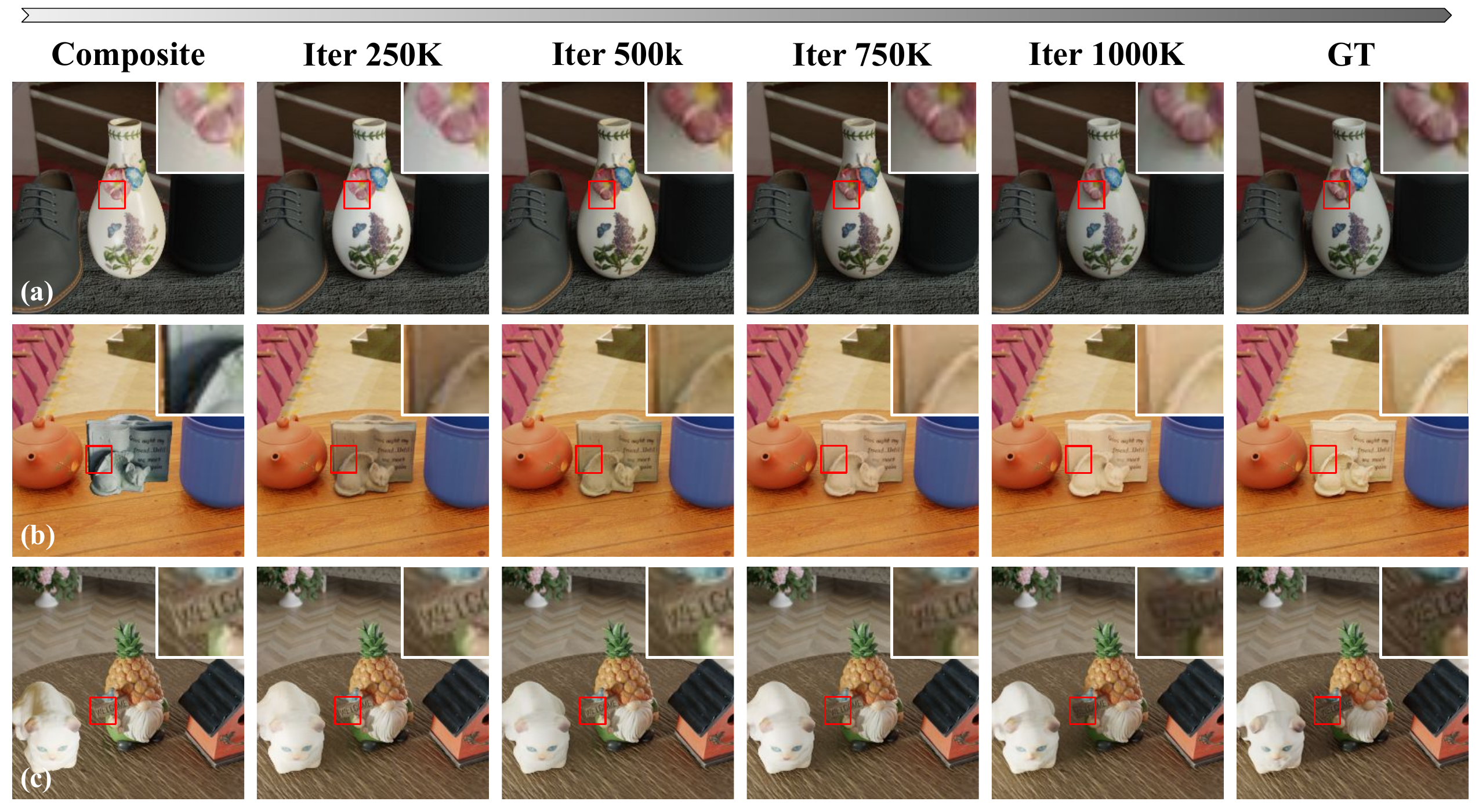}
\caption{Qualitative comparison under sequential training iterations. We demonstrate the performance of the 2D object compositing model on test set scenarios on 250K, 500K, 750K, and 1000K training iterations. At 250K iterations, the model's output diverges from the input primarily in the attenuation or removal of highlights and shadows on inserted objects. For example, the highlights on the vase surface in (a) and the shadow beneath the cat in (c) are significantly diminished. At 500K iterations, the source lighting of inserted objects is largely eliminated, while faint shadows emerge on the side opposing the scene lighting. At 750K iterations, inserted objects integrate coherently into the background scene, with generated shadows naturally cast onto the table or surrounding objects. At 1000K iterations, the model refines surface highlights and shadows with enhanced realism, emphasizing shadow details in localized regions. Notably, the carved patterns on the vase in (a) exhibit nuanced shadow variations, and the raised signboard in (c) casts a partial shadow occluded by the cat.}
\label{fig:trend}
\end{figure*}

\subsection{Learning Trend of 2D Object Compositing Model}
\label{supp:trend}

In this section, we provide a systematic analysis of the training behavior of the 2D object compositing model, which is trained for a total of 1000K iterations. As shown in Fig.~\ref{supp:trend}, the model initially learns to remove highlights and shadows from the inserted objects. In the subsequent stages, the inserted objects become progressively harmonized with the scene, and new, scene-consistent shadows and highlights are generated. In the final phase, these lighting effects are further refined to enhance realism and local detail.

\begin{figure*}[t!]
\centering
\includegraphics[width=\linewidth]{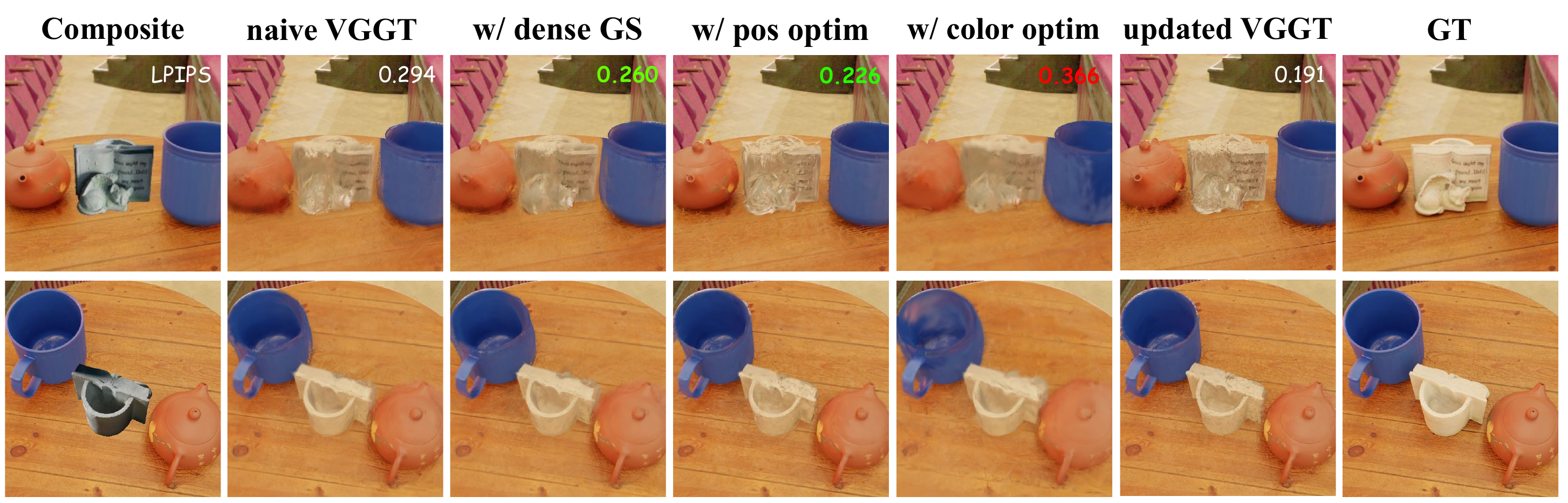}
\caption{Object insertion results based on VGGT~\cite{wang2025vggt} priors. Given multiple unposed images as input, we estimate camera poses and depth maps via VGGT to compute point cloud positions for initializing Gaussians. We conduct several comparative experiments to analyze the impact of adjustments in the Gaussian training process on the final compositing results.}
\label{fig:vggt}
\end{figure*}

\subsection{Object Compositing with VGGT Priors}
\label{supp:vggt}
VGGT has recently garnered significant attention due to its robust geometric capabilities in efficiently establishing relative relationships across unposed images. Leveraging this powerful prior, we explore the performance boundaries of our model under constrained input conditions, as shown in Fig.~\ref{fig:vggt}. When initializing Gaussians using VGGT-estimated poses and depth maps with fixed Gaussian positions and colors, the model produces blurred harmonized results. This stems from the insufficient accuracy of VGGT-estimated camera poses. To address this, we introduce positional optimization for Gaussian primitives, which enhances geometric coherence and yields sharper composited outputs. Furthermore, increasing the density of Gaussian primitives improves detail preservation. However, optimizing Gaussian colors degrades performance by inducing overfitting to training views, manifesting as noisy artifacts in the Gaussian representation and consequently deteriorating harmonization quality. Our experiments demonstrate that under VGGT-derived pose and depth constraints, denser Gaussian distributions and positional optimization positively impact final results, while color optimization adversely affects output clarity.

\subsection{Limitation and Future Work}
\label{supp:limitation}
In this paper, we propose MV-CoLight, a feed-forward model-based method for 2D-3D object compositing. While our approach achieves superior performance across diverse synthetic and real-world scenes, several limitations persist:
(a) Color bias in real-world scenes. Trained predominantly on large-scale synthetic data, the model occasionally exhibits color discrepancies when applied to certain real-world environments.
(b) Physically inconsistent illumination. Due to the absence of strict physical constraints, deviations in specular highlight positions and shadow directions may arise under complex lighting conditions.
(c) Gaussian representation limitations. Errors in Gaussian parameterization, constrained by their inherent capacity to model complex scene details, can propagate to degrade harmonization quality.
To address these problems, future work may focus on:
(a) Integration of physical constraints. Enhancing shadow consistency by estimating light source positions and constraining shadow regions.
(b) 4D scene harmonization. Extending harmonization to 4D dynamic scenes to enable consistent movement of inserted objects within dynamic environments.

\end{document}